\documentclass[10pt, journal]{IEEEtran}

\usepackage[table]{xcolor} 
\usepackage{array}

\IEEEoverridecommandlockouts
\usepackage{tabularx}
\usepackage{multicol}
\usepackage{cite}
\usepackage{amsmath,amssymb,amsfonts}
\usepackage{algorithmic}

\usepackage{graphicx}
\usepackage{subfigure}
\usepackage{textcomp}
\usepackage{soul}
\usepackage{xcolor}
\usepackage{url}
\def\BibTeX{{\rm B\kern-.05em{\sc i\kern-.025em b}\kern-.08em
    T\kern-.1667em\lower.7ex\hbox{E}\kern-.125emX}}
    
\usepackage[linesnumbered,ruled,vlined]{algorithm2e}
\usepackage{algorithmic}

\usepackage{pifont}
\newcommand{\cmark}{\ding{51}}%
\newcommand{\xmark}{\ding{55}}%
\usepackage{makecell}

\makeatletter
\def\ps@IEEEtitlepagestyle{%
  \def\@oddhead{%
    \hfill
    \begin{minipage}{0.9\textwidth}
      \centering\footnotesize
      \textcopyright\ IEEE. This is the author's accepted manuscript of a paper
      accepted for publication in \emph{IEEE Transactions on Network and Service Management}.
      The final published version is available at IEEE Xplore: 10.1109/TNSM.2024.3447532
    \end{minipage}
    \hfill
  }
  \def\@oddfoot{}
}
\makeatother

\begin{document}

\title{Real-Time Adaptive Anomaly Detection in Industrial IoT Environments}

\author{Mahsa Raeiszadeh$^{\dagger}$, Amin Ebrahimzadeh$^{\dagger}$, Roch H. Glitho$^{\dagger**}$,~\emph{Senior~Member,~IEEE}, \\ Johan Eker$^{*}$, and Raquel A. F. Mini$^{*}$
        \\ $^{\dagger}$CIISE, Concordia University, Montr\'eal, QC, Canada\\
$^{*}$Ericsson Research, Lund, Sweden\\
$^{**}$Computer Science Programme, University of Western Cape, Capetown, South Africa}

\maketitle

\begin{abstract}

To ensure reliability and service availability, next-generation networks are expected to rely on automated anomaly detection systems powered by advanced machine learning methods with the capability of handling multi-dimensional data. Such multi-dimensional, heterogeneous data occurs mostly in today's Industrial Internet of Things (IIoT), where real-time detection of anomalies is critical to prevent impending failures and resolve them in a timely manner. \textcolor{black}{ However, existing anomaly detection methods often fall short of effectively coping with the complexity and dynamism of multi-dimensional data streams in IIoT.} In this paper, we propose an adaptive method for detecting anomalies in IIoT streaming data utilizing a multi-source prediction model and concept drift adaptation. The proposed anomaly detection algorithm merges a prediction model into a novel drift adaptation method resulting in accurate and efficient anomaly detection that exhibits improved scalability. Our trace-driven evaluations indicate that the proposed method outperforms the state-of-the-art anomaly detection methods by achieving up to an 89.71\% accuracy (in terms of Area under the Curve (AUC)) while meeting the given efficiency and scalability requirements.
\end{abstract}

\begin{IEEEkeywords}
Anomaly Detection, Real-time Analytics, Concept Drift, Streaming Data, Industrial Internet of Things (IIoT)
\end{IEEEkeywords}

\section{Introduction}
\label{sec:intro}

Streaming data from the Industrial Internet of Things (IIoT) and time-sensitive applications are highly dynamic, heterogeneous, and large-scale, demanding rapid and effective real-time analytics~\cite{Putina2021}. It is critical to maintain the robustness and reliability of time-sensitive systems with respect to unexpected hardware and/or software failures, particularly when the system scales up to a huge cluster size~\cite{Dromard2017}. 
Thus, practical methods for identifying anomalies in a bounded time to avoid severe losses are more crucial than ever, as anomalies in time-sensitive services can not only have detrimental impacts on the overall performance but also incur excessive maintenance costs. 
It is expected that time-sensitive services and applications like autonomous driving and smart manufacturing will drive the evolution of next-generation mobile networks~\cite{jiang2021road}. To keep up with the stringent requirements of time-sensitive applications, next-generation networks must be trustworthy and capable of self-maintenance with regard to their reliability and availability. High complexity and increased density of mobile networks mandate the need for automated network management with the capability of detecting anomalies while adapting to environment changes~\cite{dealmeida2021abnormal}. This gives way to the so-called self-healing systems, where the dynamic and complex characteristics of the network provide a diverse array of measurements and metrics to be translated into valuable knowledge to support anomaly detection in real time. \textcolor{black}{Furthermore, given their ability to autonomously rectify anomalies and optimize operations without human intervention, the adoption of self-healing systems~\cite{dealmeida2021abnormal} within IIoT environments is essential for enhancing operational reliability and efficiency in complex production scenarios.}

Automating anomaly detection in IIoT has become increasingly critical and challenging due to several reasons. First, IIoT networks typically generate high-velocity and high-volume data, which means that many IIoT devices generate data with high frequency, and due to the growing number of IIoT devices, large amounts of IIoT data are generated worldwide \cite{yang2022iot}. To cope with the high volume and high velocity of the data, the processing module should stick to the one-pass criterion and needs to be processed upon their arrival in real time.
Second, depending on its intrinsic characteristics, time-series data may be periodic, seasonal, and/or irregular and possess a high correlation. Therefore, it is not legitimate to assume that time-series data is independent and identically distributed; different features of the sample are correlated to each other
\cite{Ahmad2017}. Besides, in real-world scenarios, unknown anomalous events may also occur in IIoT systems due to unavoidable changes in the statistical characteristics of the data, commonly referred to as concept drift~\cite{yang2022iot}. Third, In order to maintain robustness and accuracy in time-sensitive applications, it is crucial to detect anomalous events in real time. \textcolor{black}{Some works have focused on anomaly detection in smart grids, employing various methods to monitor and address irregularities in power consumption and grid operation~\cite{banik2023anomaly}. We note that anomaly detection in the IIoT environments focuses on challenges unique to high-velocity, high-volume, and multi-dimensional data of industrial systems, with an emphasis on real-time processing and adaptation to concept drift, which are crucial for operational efficiency in a wide range of industrial applications beyond the energy sector.
}

\begin{figure*}[!t]
\centerline{\includegraphics[width=2.1\columnwidth,keepaspectratio]{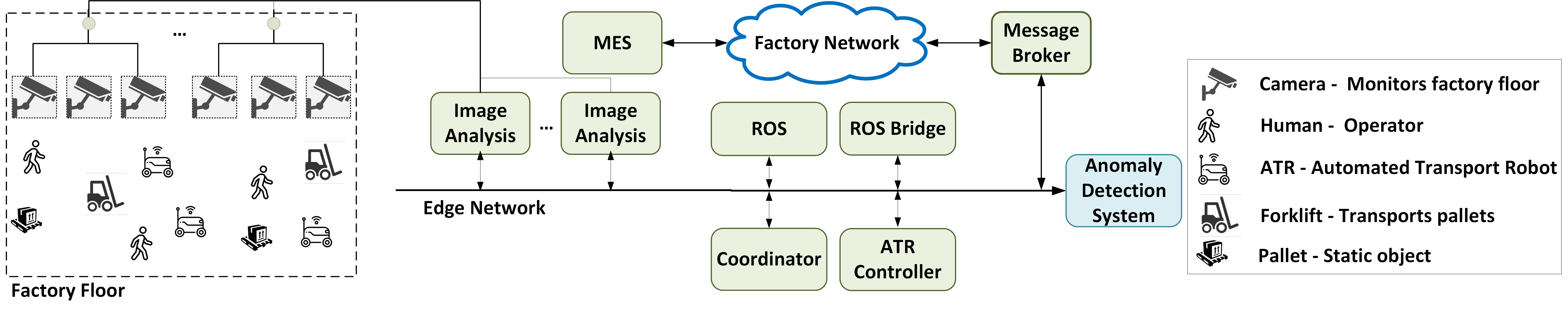}}
\caption{\textcolor{black}{Illustration of an IIoT-based smart manufacturing environment as the motivating scenario.}}
\label{fig:1}
\end{figure*}

Recently, in \cite{Raeiszadeh2023} we have proposed a prediction-based approach for the automated detection of application-level anomalies in time-series data. The proposed solution comprises a multi-source prediction module using a Long-Short Term Memory (LSTM) based deep learning and a novel detection method, which utilizes the prediction of future sequences to determine whether or not an arriving data is
anomalous. To enhance the accuracy of the prediction model, we have devised a weighting mechanism that considers the ensemble of consecutive predictions, turning it into a chronological procedure to enable real-time detection. While the work \cite{Raeiszadeh2023} addresses the accuracy and real-time aspects of the problem, the concept drift and scalability with the number of dimensions were not considered.

In this work, we build on \cite{Raeiszadeh2023} and address the scalability requirement of real-time anomaly detection and the need for drift detection and adaptation while considering the real-time requirements. By combining the window technique and model update to the Prediction-Driven Anomaly Detection (PDAD-SID) method proposed in \cite{Raeiszadeh2023}, we propose a drift-adapted, real-time anomaly detection algorithm called Scalable and Adaptable Prediction Driven Anomaly Detection (SAPDAD). The proposed method exploits the PDAD-SID method for multi-source prediction to estimate anomaly probabilities along with a Genetic Algorithm (GA) based model optimization. Further, for concept drift detection and adaptation, we have proposed our so-called Real-Time Optimized Adaptive Windowing (RealTimeOAW), which regularly updates the anomaly detection model to adapt to concept drifts. The proposed SAPDAD method addresses the high velocity and volume of data streams, as it uses an adaptive sliding window to store and process the latest records only once. Further, it leverages the Principal Component Analysis (PCA) to reduce the dimensions and incorporate correlations between different features, thus increasing detection efficiency as well as scalability. The key contributions of this paper are summarized as follows: 

 \begin{itemize}
%
    \item We propose an optimized and scalable anomaly detection method that exploits a multi-source prediction model equipped with LSTM to detect anomalous events in real time via a parallel calculation approach, which scales well with the number of dimensions.
    \item We propose a drift adaptation method called RealTimeOAW to tackle the challenge of concept drift by regularly updating the anomaly detection model to adapt to heterogeneous environments while considering time and accuracy requirements. Our proposed RealtimeOAW incorporates sliding and adaptive window-based methods with a performance-based approach to adapt to concept drifts.
    \item We have carried out trace-driven evaluations using three real-world datasets, namely, KDDCup99\footnote{\url{https://datahub.io/machine-learning/kddcup99}} \cite{spinosa2008cluster}, IoTID20\footnote{\url{https://sites.google.com/view/iot-network-intrusion-dataset/home}} \cite{ullah2020scheme}, and WUSTL-IIoT~\cite{zolanvari2019machine} in terms of detection accuracy, processing rate, and computational efficiency.
    
\end{itemize}

The remainder of this paper is structured as follows. Section~\ref{relatedwork} presents the motivating scenario, requirements, and related work. The system model
is presented in Section~\ref{system}. Section~\ref{solution} describes our proposed method. Our evaluation results are presented in Section~\ref{evaluation}. Section~\ref{conclusion} concludes the paper.
\section{Motivating Scenario, Requirements, and Related Work}
\label{relatedwork}

In this section, we first present an illustrative example as our motivating scenario, which is then used to derive the key requirements. We then review the existing works.

\subsection{Motivating Scenario and Requirements}

Figure~\ref{fig:1} illustrates an IIoT smart manufacturing environment, which serves as an example to motivate our research problem. As shown in Fig.~\ref{fig:1}, camera-based positioning is used to control the Automated Transport Robots (ATRs) on the factory floor. Cameras covering the factory floor are connected via a Local Area Network (LAN). To avoid any blind spots, neighboring cameras have overlapping coverage. Several cameras are mounted as visual sensors to capture video from the factory floor. The captured video streams are processed to locate and identify mobile objects and/or obstacles, building a world model that can be used for path planning and obstacle avoidance. The planned paths are sent in segments to the ATRs by the ATR Controller. The Robot Operating System (ROS) is used for robot control. Given that both ROS v.1 and ROS v.2 nodes coexist, ROS Bridge is used to translate from ROS v.1 to ROS v.2. A number of tasks need to be accomplished partly in parallel, like scheduling the fleet of ATRs and the Manufacturing Execution System (MES). One observation of broker-based messaging systems is that they support some of the service meshes' functionality. For example, they can provide basic load balancing using their queue and consumer groups. The smart manufacturing system has a latency requirement to control ATRs. ATRs may show anomalous behaviors, such as being on the wrong path or moving without any updated path. 

In such scenarios, the collected data is typically multi-dimensional streaming time-series data, which can be utilized to detect the anomalous behaviors of ATRs in real time. However, in the example described above, two challenges arise. First, factory environments are highly dynamic, resulting in the continuous exposure of streaming data to various forms of concept drift. For example, sensing components may be replaced/updated periodically (e.g., due to aging) or the events monitored by them may change over time, or IoT event factors, such as system updates, replacement of IoT devices, and anomalous network events, may happen. Concept drifts can cause degradation of the detection model, making real-time anomaly detection more challenging. Second, the factory environment generates high-dimensional time-series data, which can be irrelevant, noisy, and/or redundant. It is, therefore, difficult to resolve the dilemma of accuracy and scalability in detecting anomalies in a factory environment with time constraints. Clearly, the smart factory example explained above requires an anomaly detection system that is (i) real-time, (ii) concept drift adapted, (iii) accurate, (iv) time efficient, and (v) scalable.


\subsection{Related Work}
\label{sec:related_work}

In the following, we review the related work on anomaly detection followed by the existing concept drift adaptation methods. 

\subsubsection{Anomaly Detection}

The existing anomaly detection methods can be classified into three categories of (a)~statistical, (b)~data-driven, and (c)~deep learning methods.

\paragraph{Statistical Methods}
 
Statistical anomaly detection primarily uses distance-based methods, which calculate anomaly scores through k-nearest-neighbors distance~\cite{quek2019iot}, local outlier factor (LOF)\cite{xie2022abnormal}, or histogram-based outlier scores\cite{paulauskas2019application}. Common models include the autoregressive model, moving average, and autoregressive moving average models~\cite{hao2021hybrid}. While effective for consistent trends, these models are noise-sensitive and assume known data distribution, which is often not valid for high-dimensional datasets with unpredictable events. Histogram-based methods are inadequate for high-dimensional data due to their inability to analyze feature interactions. Additionally, selecting the optimal k value for accurate LOF computation in the LOF algorithm~\cite{xie2022abnormal} is challenging.

\paragraph{Data-Driven Methods}

Data-driven techniques in machine learning can be supervised, unsupervised, or semi-supervised \cite{nassif2021machine}. Supervised learning \cite{laptev2015,Liu2015} involves training models with labeled data, which typically requires human input. This, however, can be time-consuming and sometimes infeasible in practical scenarios, where some anomalies may be unknown. \textcolor{black}{
Aman et al.~\cite{aman2022machine} developed a machine learning-based method to secure IoT devices by analyzing their memory dumps. This technique identifies firmware tampering and maintains internal integrity. They addressed multiple attack vectors in IoT environments, such as firmware modification attacks altering IoT device firmware, exploitation of firmware vulnerabilities, and botnet formations using compromised devices for coordinated attacks. Javaid et al.~\cite{javaid2020defining} proposed a blockchain-based remote attestation protocol for IoT environments, utilizing a secure blockchain for device registration and Physical Unclonable Functions (PUF) for attestation.}

Alternatively, unsupervised learning methods for anomaly detection typically aim to identify the unique features, which help distinguish abnormal data from normal data. To this end, iForest~\cite{li2021similarity} uses a forest of isolation trees from training samples to compute anomaly scores based on path lengths. iForest is efficient for large and high-dimensional data, though its detection performance reaches convergence rapidly with a limited number of trees. In~\cite{Putina2021}, an unsupervised clustering-based method called Online Anomaly Detection in Data Streams (ODS) was introduced for real-time anomaly detection in telemetry data. However, it may easily be trapped in local minima and do not behave proactively when changes occur.

\paragraph{Deep Learning Methods}
\textcolor{black}{Many works utilize deep learning for anomaly detection~\cite{kauhsik2023iot}}.
Munir et al.~\cite{Munir2019} introduced DeepAnT, which relies on deep Convolutional Neural Network (CNN). This method is effective for anomaly detection in time-series data, even with small datasets. In~\cite{Zhao2021} a multilayer convolutional recurrent autoencoder is developed for detecting anomalies in multivariate time series. Zhou et al.\cite{Zhou2021} developed a variational LSTM (VLSTM) model for unbalanced, high-dimensional industrial data. The authors of \cite{WuLSTM2020} proposed an LSTM-Gauss-NBayes method, which combines LSTM-NN and Naive Bayes for anomaly detection in IIoT. A real-time anomaly detection method for NoSQL systems called RADAR was proposed in \cite{Chouliaras2020}, which identifies anomalous events by extracting process information during resource monitoring. Malhorta et al.~\cite{malhotra2015} used a multiple-prediction technique assessing anomalies through prediction error and Gaussian error distribution. Wang et al.~\cite{wang2022improved} proposed an improved LSTM method to detect anomalies in time series data by forecasting and comparing predicted and observed sequences, addressing the issue of data with diverse distributions and the lack of labeled anomalous data.
\textcolor{black}{Mothukuri et al.~\cite{mothukuri2021federated} proposed an anomaly detection method using Gated Recurrent Units (GRUs) models for the real-time and proactive identification of network intrusions in IoT systems.}
\textcolor{black}{Nandanwar and Katarya~\cite{nandanwar2024deep} proposed a transfer learning model called TL-BILSTM IoT, which is designed to predict intrusion detection in IIoT environments. Their model combines CNN and Bidirectional LSTM to detect and classify botnet attacks.}

We note that most of the existing deep learning techniques can not be directly applied to IIoT scenarios because (1) they fail to adapt to frequently changing time-series data, (2) data is usually not available for training due to privacy concerns, and (3) they demand substantial computing power and extended training time.

 \begin{table*}[!t]
\begin{center}
\caption{Comparison of existing anomaly detection techniques with this work.}
\label{tablecompare}
\begin{tabular}{|l|lllll|}
\hline
\multicolumn{1}{|c|}{Related Work} & \multicolumn{5}{c|}{Requirements}                                                                                                                                                                 \\ \cline{2-6} 
\multicolumn{1}{|c|}{}                              & \multicolumn{1}{l|}{Real-Time Anomaly Detection} & \multicolumn{1}{l|}{Concept Drift Adaptation} & \multicolumn{1}{l|}{Accuracy} & \multicolumn{1}{l|}{Time-Efficiency} & Scalability \\ \hline

ODS \cite{Putina2021}& \multicolumn{1}{c|}{\cmark}                          & \multicolumn{1}{c|}{\xmark}                                        & \multicolumn{1}{c|}{\xmark}           & \multicolumn{1}{c|}{\xmark}            &        \multicolumn{1}{c|}{\xmark}      \\ \hline
ORUNADA \cite{Dromard2017}                                                 & \multicolumn{1}{c|}{\cmark}                         &\multicolumn{1}{c|}{\cmark}                                         & \multicolumn{1}{c|}{\xmark}            & \multicolumn{1}{c|}{\xmark}            &       \multicolumn{1}{c|}{\xmark}         \\ \hline

OLINDDA\cite{spinosa2008cluster}                                                   & \multicolumn{1}{c|}{\cmark}                            & \multicolumn{1}{c|}{\cmark}                                       & \multicolumn{1}{c|}{\cmark}            & \multicolumn{1}{c|}{\xmark}           &         \multicolumn{1}{c|}{\xmark}     \\ \hline
iForest\cite{li2021similarity}                                             & \multicolumn{1}{c|}{\xmark}                         &\multicolumn{1}{c|}{\xmark}                                         & \multicolumn{1}{c|}{\cmark}            & \multicolumn{1}{c|}{\xmark}            &       \multicolumn{1}{c|}{\xmark}              \\ \hline
DeepAnt\cite{Munir2019}                                           & \multicolumn{1}{c|}{\xmark}                            & \multicolumn{1}{c|}{\xmark}                                       & \multicolumn{1}{c|}{\cmark}            & \multicolumn{1}{c|}{\xmark}           &  \multicolumn{1}{c|}{\xmark}           \\ \hline
RADAR\cite{Chouliaras2020}                                           & \multicolumn{1}{c|}{\cmark}                          & \multicolumn{1}{c|}{\xmark}                                           &\multicolumn{1}{c|}{\xmark}          & \multicolumn{1}{c|}{\xmark}           &      \multicolumn{1}{c|}{\xmark}        \\ \hline

\cite{yu2015scalable}                                             & \multicolumn{1}{c|}{\cmark}                         &\multicolumn{1}{c|}{\cmark}                                         & \multicolumn{1}{c|}{\xmark}            & \multicolumn{1}{c|}{\xmark}            &       \multicolumn{1}{c|}{\xmark}           \\ \hline
A-Detection\cite{wang2021concept}                                          & \multicolumn{1}{c|}{\cmark}                         &\multicolumn{1}{c|}{\cmark}                                         & \multicolumn{1}{c|}{\cmark}            & \multicolumn{1}{c|}{\xmark}            &       \multicolumn{1}{c|}{\xmark} \\ \hline

ASTREAM \cite{yang2022astream}                                                & \multicolumn{1}{c|}{\cmark}                         &\multicolumn{1}{c|}{\cmark}                                         & \multicolumn{1}{c|}{\cmark}            & \multicolumn{1}{c|}{\cmark}            &       \multicolumn{1}{c|}{\xmark}         \\ \hline
Online RNN-AD \cite{saurav2018online}                                                & \multicolumn{1}{c|}{\xmark}                        & \multicolumn{1}{c|}{\cmark}                                      & \multicolumn{1}{c|}{\cmark}        & \multicolumn{1}{c|}{\cmark}      & \multicolumn{1}{c|}{\xmark}            \\ \hline
PDAD-SID  \cite{Raeiszadeh2023}                                             & \multicolumn{1}{c|}{\cmark}                            & \multicolumn{1}{c|}{\xmark}                                       & \multicolumn{1}{c|}{\cmark}            & \multicolumn{1}{c|}{\cmark}           &       \multicolumn{1}{c|}{\xmark}       \\ \hline
This work                                              & \multicolumn{1}{c|}{\cmark}                            & \multicolumn{1}{c|}{\cmark}                                       & \multicolumn{1}{c|}{\cmark}            & \multicolumn{1}{c|}{\cmark}           &         \multicolumn{1}{c|}{\cmark}    \\ \hline
\end{tabular}

\end{center}
\end{table*}

\subsubsection{Anomaly Detection with Concept Drift Adaptation}

Some studies have studied the problem of anomaly detection in the presence of concept drift. Dromard et al. \cite{Dromard2017} presented ORUNADA, an unsupervised network anomaly detection algorithm using incremental grid clustering and a discrete sliding window to update features. 
%
Spinosa et al. \cite{spinosa2008cluster} introduced OLINDDA, an unsupervised clustering algorithm for detecting anomalies in streaming data by modeling normal data with a hypersphere around normal data clusters. 
Yu et al. \cite{yu2015scalable} developed a framework for detecting anomalies in large-scale datasets, clustering nodes based on geographic location and network topology, and employing a two-phase majority voting algorithm to identify anomalous nodes. This method effectively detects new anomalies by spotting deviations from majority behavior, but its adaptability to system behavior changes during runtime is unclear.

In \cite{wang2021concept}, the so-called A-Detection method was proposed, which uses reservoir sampling and singular value decomposition (SVD) to analyze data streams and applies Jensen Shannon (JS) divergence to detect anomalies. Yang et al. \cite{yang2022astream} introduced ASTREAM, an anomaly detection system for data streams in real-time scenarios. ASTREAM incorporates sliding windows, concept drift detection, and updating models in a hashing-based locality-sensitive iForest model to address the challenges of continuous data streams.
\textcolor{black}{
Presenting a comprehensive performance evaluation using the KDDCup99 dataset, the authors of \cite{yang2022astream} have shown that the ASTREAM algorithm is robust in handling the challenges of IIoT data streams, such as managing infinite data and adapting to data distribution changes. We note, however, that this work mainly addresses intrusion detection, thus having a narrower scope compared to our work, where we consider a broader range of dynamic IIoT environments and use various datasets to demonstrate wider applicability. 
%
%
%
}

In \cite{saurav2018online}, a deep RNN-based method called Online RNN-AD was introduced for online time-series anomaly detection. This method incorporates local normalization of incoming data and incremental neural network retraining, showing adaptability to concept drift in time-series data. 
%
%
In \cite{ippoliti2016online}, an online and adaptive anomaly detection model was proposed by employing a one-class SVM, which uses unlabeled data to create a hyperplane, isolating a region with most normal vectors, and evaluates anomalies based on their proximity to this hyperplane. 
This method allows dynamic input normalization and adaptation based on operator feedback, offering a confidence level for each anomaly. 
We note that recalculating the hyperplane for conflicting operator feedback can hamper real-time accuracy. Besides, the SVM approach suffers from the curse of dimensionality and struggles with large feature sets.

Table~\ref{tablecompare} compares the existing relevant works with this paper based on our requirements. Existing studies 
simply detect anomalies in streaming data and they
mostly ignore real-time detection; for example, some works, e.g.,~\cite{li2021similarity},~\cite{Munir2019}, store the observed data for processing and do not stick to the one-pass criterion. Therefore, they fail
to deliver meaningful results in a timely manner.  Furthermore, some works (e.g.~\cite{Raeiszadeh2023},~\cite{Putina2021},~\cite{Munir2019},~\cite{li2021similarity}) fail to incorporate concept drift adaptation, crucial for addressing the changing statistical behavior of IIoT data streams. Moreover, none of these studies offer scalable solutions for handling high-dimensional IIoT time-series data. 

\section{System Model and Problem Statement}
\label{system}

\subsection{System Model}\label{system-sub}
\begin{figure}[!t]
\centerline{\includegraphics[width=9cm,keepaspectratio]{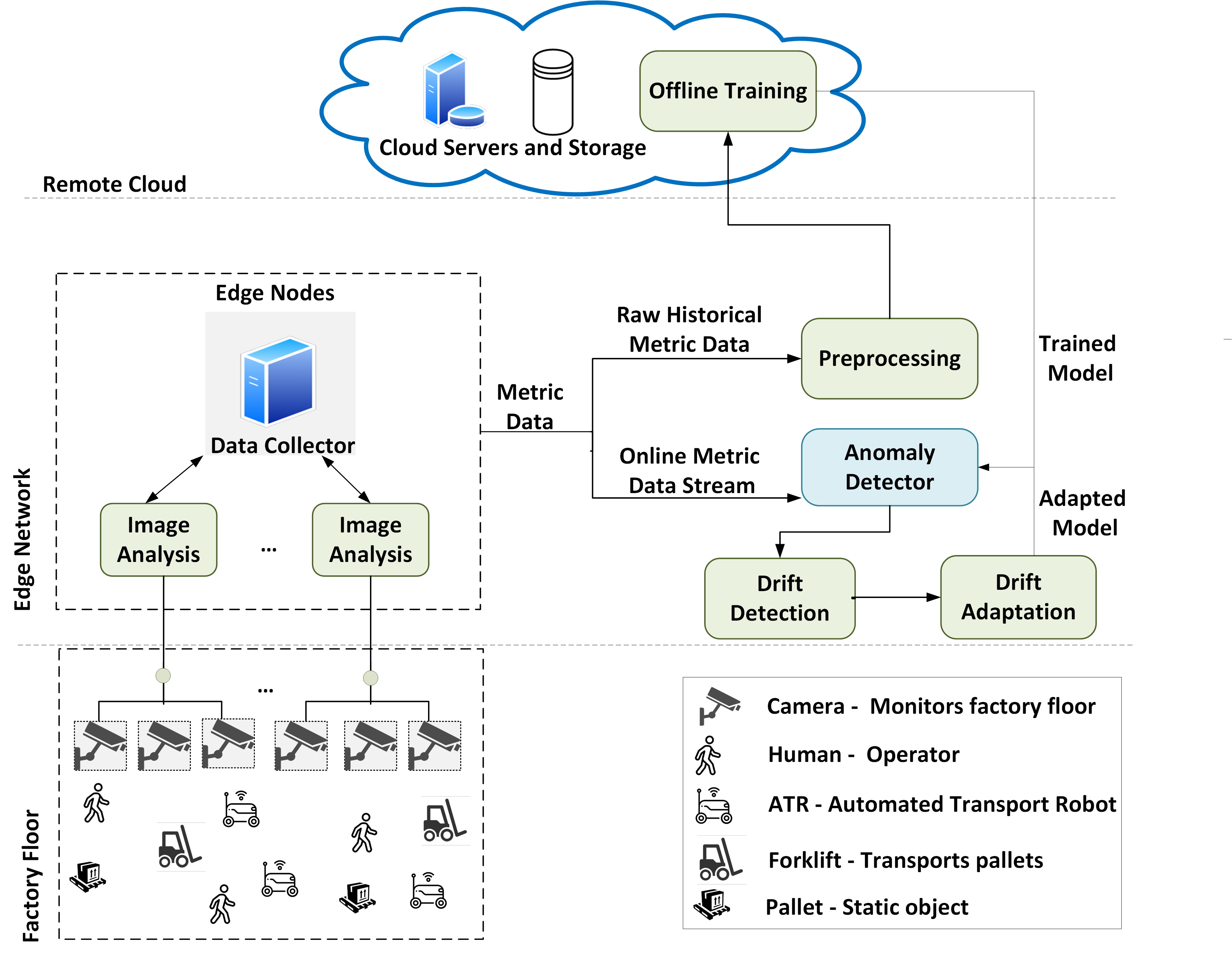}}
\caption{\textcolor{black}{System model.}}
\label{fig:system_model}
\end{figure}

Figure~\ref{fig:system_model} illustrates a smart factory environment, which comprises a factory floor, an edge network, and a remote cloud. The factory floor comprises human operators, dynamic objects such as ATRs, forklifts, and static obstacles like pallets. ATRs are assisted by visual object detection in order to pick up the components without human assistance and deliver them to the production line. Multiple cameras, each equipped with an Image Analysis module, are mounted in the area. Each camera transmits the captured video to the Image Analysis module via REST API, which is typically used in industrial systems for inter-component communication~\cite{domenech2016providing}. The captured video streams are analyzed in the Image Analysis module to detect, classify, identify, position, and track the objects on the factory floor to build a world model that can be used for objects' traveling, scheduling, and obstruction avoidance. For instance, scheduling ATRs' travel paths involves planning how ATRs should travel to arrive at the destination at a certain time and is based on a set of objects with predicted trajectories. 

Almost all of the processing, such as travel planning and scheduling, is carried out in the edge network rather than the devices on the factory floor. As shown in Fig.~\ref{fig:system_model}, the data collector in the edge network has the crucial task of collecting the data in the form of IIoT time-series metric data. The time-series metric data can be either one-dimensional, where a sequence of measurements from the same features are collected, or multi-dimensional, where a sequence of measurements from multiple features or sensors are collected. The smart factory system may generate multi-dimensional series metric data with intricate features such as ATRs status, which is stationary or moving, or the wheel velocity of ATRs. The raw collected metric data is passed to the Preprocessing module to clean and normalize. The preprocessed data undergoes offline training to develop a time-series prediction model. Subsequently, the trained model is deployed in the Anomaly Detector system, which is responsible for detecting application-level anomalies. Moreover, the data collector also receives an online metric data stream and feeds it to the Anomaly Detector in order to carry out a real-time detection of anomalies.

 Ideally, detecting anomalies should be as simple as computing the distance between the ground truth (i.e., class labels) and the target (i.e., record). A record is a data sample at a specific moment within a series. A record can take the form of a vector within a multi-dimensional time-series data or a scalar value within a one-dimensional time-series data. However, data labels may not always be available in practice since labeling the data is often challenging, mainly because anomalous records (i.e., records that do not conform to normal or expected records), are rare compared to normal records. To cope with this, a prediction-driven anomaly detection mechanism is needed, where the ground truth is replaced by their predicted values, which can be generated by a time-series prediction algorithm. Clearly, such an approach is dependent on the accuracy of the prediction module, which is usually affected by the anomalous records in the historical data.

Due to the dynamic nature of configurations and workloads caused by, for instance, updates to ROS or replacement of ATRs, the distribution of incoming data may undergo frequent changes over time, resulting in concept drift. This, in turn, leads to false anomaly detection and results in poor performance of the detection model. To address this issue, the Drift Detector within the edge network takes in the anomaly rate (which specifies the likelihood that an anomaly has occurred) as the input. If the anomaly rate consistently increases, the Drift Detector identifies a concept drift and triggers an alert to indicate drift occurrence. Following a drift alert, the system collects new data, and the Drift Adapter in the edge network adjusts the anomaly detection model to adapt to the concept drift using the newly received data.
\textcolor{black}{
In the smart factory environment, anomaly detection is crucial for addressing a wide range of cybersecurity and physical vulnerabilities, particularly in industrial control systems. Key concerns include data interception and tampering, where IIoT data streams are vulnerable to manipulation by external attackers, potentially hiding real anomalies or creating false ones. \textcolor{black}{This risk extends to unauthorized access and control over critical IIoT systems, leading to failures such as operational disruptions and data breaches, which often result from root causes such as software vulnerabilities, insecure network configurations, and/or compromised third-party services.}
Network-based threats such as DDoS attacks disrupt services and pose safety hazards, representing another failure mode. Insider threats involving system users compromising IIoT security and physical security breaches, such as tampering with IIoT devices, are also key components of the threat landscape. \textcolor{black}{Also, failures like application crashes due to root causes such as software bugs or resource leaks can cause significant operational downtime. Similarly, failures such as network latency, commonly caused by root causes like bandwidth overload or hardware malfunctions, can significantly disrupt manufacturing operations.}
}
\textcolor{black}{
In this paper, we make the following assumptions:
\begin{itemize}
 \item We assume that the smart manufacturing environment is equipped with automated network management, such as a self-healing system~\cite{dealmeida2021abnormal}.
    \item We assume that the data collected by different types of industrial sensors on the factory floor is transmitted to edge nodes through REST API, which is typically used in industrial systems for inter-component communication~\cite{domenech2016providing}.
    \item We assume that IoT devices generate a high-volume data stream with high frequency that is not independent and identically distributed (non-i.i.d.) and possesses high correlation. Additionally, this data varies in format and characteristics, often including irrelevant, noisy, or redundant elements~\cite{Ahmad2017}.
    \item We assume that data patterns change over time, due to a phenomenon commonly known as concept drift, which can manifest in various forms. We specifically consider three types of concept drifts, namely, sudden, gradual, and recurring~\cite{yang2022iot}.  
    \item We assume that various types of anomalies may happen in the smart manufacturing environment. These include hardware anomalies (e.g., caused by environmental interference, device malfunction, or reading errors) and software anomalies (e.g., resulting from program exceptions, transmission errors, or malicious attacks)~\cite{dealmeida2021abnormal}.
\end{itemize}
}

\subsection{Problem Statement}
\label{problem}

In IIoT environments, a significant portion of the data generated consists of time-series data exhibiting temporal correlations, where a record collected at a one-time point may have connections to previously collected data records. Time-series data consist of successive observations collected in chronological order at each time slot $t$. In our study, we define the stream of records as $R=\left\{\mathbf{r}_{0}, \mathbf{r}_{1}, ..., \mathbf{r}_{M}\right\}$, $\left(\mathbf{r}_{i} \in \mathbb{R}, i \in[0, M]\right)$, from $M$ time instants and each data record $\mathbf{r}_{i}$ is an $D$-dimension vector. The status of the system at time slot $t$ is determined by comparing the actual value $y_i$ and the predicted value $\hat{y}_i$. Thus, data stream $R$ is a sequence of observations $\mathbf{r}_{i}=\left(\boldsymbol{x_i}, y_i\right)$, where $\boldsymbol{x_i}$ is the vector of features, and $y_i$ is the actual (independent) output of the prediction problem.

Supervised learning algorithms try to learn from past data to identify the relationship between the features and the target output. In the case of prediction, the learned model ${F}\left(\boldsymbol{x_i}\right) \mapsto y_i$ is used to predict $\hat{y}_i$ for the upcoming observations. 
In this non-stationary environment, the distribution
of input data frequently undergoes changes, leading to the occurrence of concept drift.
 At time slot $t$, the observation is generated by a source with a joint probability distribution $P_t(\boldsymbol{x_i}, y_i)$. The data distribution is considered stationary if all records follow the same distribution. Concept drift occurs if for two distinct points in time slots $t_{1}$ and $t_{2}$, there exits $\boldsymbol{x_i}$ such that $P_{t_{1}}(\boldsymbol{x_i}, y_i) \neq P_{{t_{2}}}(\boldsymbol{x_i}, y_i)$.
The joint probability $P_{t}(\boldsymbol{x_i}, y_i)$ can be calculated by:
\begin{equation}
P_{t}(\boldsymbol{x_i}, y_i)=P_{t}(\boldsymbol{x_i}) \times P_{t}(y_i \mid \boldsymbol{x_i}),
\end{equation}
where $P_{t}(\boldsymbol{x_i})$ denotes the marginal probability and $P_{t}(y_i \mid \boldsymbol{x_i})$ represents the posterior probability.
At each time slot $t$, the objective is to drive a model $F_{t}$ for $P_t(\boldsymbol{x_i}, y_i)$, which can be expressed as:

\begin{equation}
F_{t}=\underset{f \in \mathcal{H}}{\arg \min } \mathbb{E}_{(\boldsymbol{x}, y) \in P_t(\boldsymbol{x}, y)}[\ell(\hat{y}, y)],
\end{equation}
where $\mathcal{H}$ represents the set of hypotheses, $\mathbb{E}(\cdot)$ indicates the expected value of the record's features, and $\ell(\cdot, \cdot)$ represents the loss function, which can be defined as a Sequence Inconsistency Distance (SID) metric, which is explained in technically greater detail in Section~\ref{realtimephase}.


\begin{table}[t]
\caption{summary of main notations}
\label{notation}
\begin{tabular}{ll}
\hline
Symbol                         &  Description                        \\ \hline
R                 & A stream of records                  \\
$\mathbf{r}_{i}$                          &  An actual record                            \\
$\mathbf{\hat{r}}_{i}$                          &  A predicted record                            \\
D                &  Number of features (Dimensions)               \\
$\mathbf{r}^{(l)}_{i}$                & The component l of record i               \\
N                     &Length of sequence                    \\
$\mathbf{S}^{N}_{i}$                & Sequence of N consecutive actual records ending at record i                 \\
$\mathbf{\hat{S}}^{N}_{i}$                & Sequence of N consecutive predicted records ending at record i                 \\
M &  Number of records in the dataset \\
L                     & Length of prediction horizon                        \\
$P(i)$                     & Probability of record i being normal                     
\\
$T$                     &Anomaly probability threshold                    
\\
 \hline
\end{tabular}
\end{table}

\begin{figure*}[!t]
\centerline{\includegraphics[width=2\columnwidth,keepaspectratio]{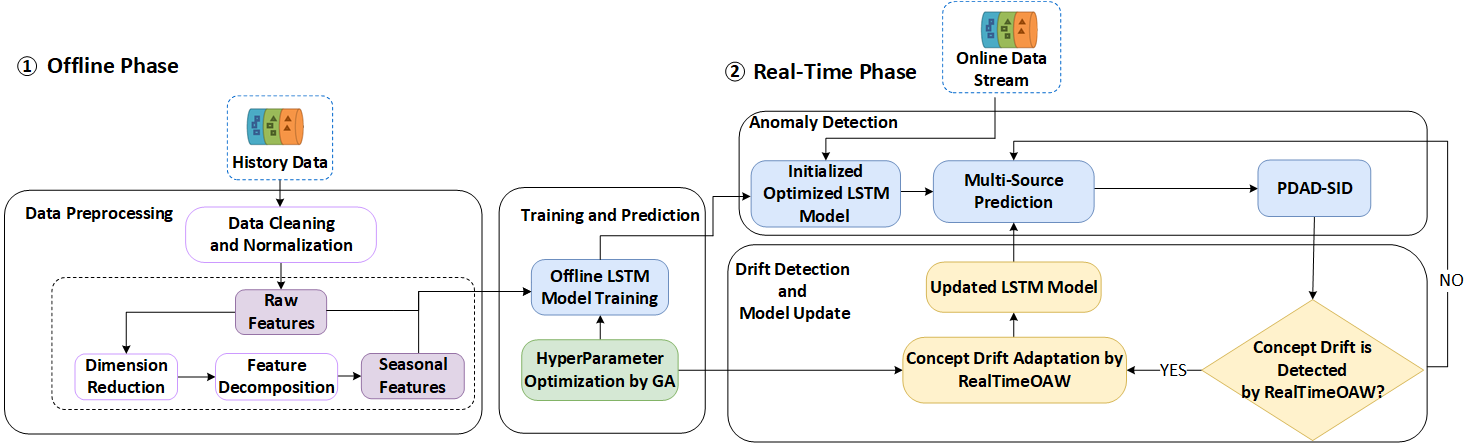}}
\caption{Overview of the proposed Scalable and Adaptable Prediction-Driven Anomaly Detection (SAPDAD).}
\label{fig1}
\end{figure*}

In the environment described in Section~\ref{system-sub}, considering the diversity of architectures employed in the edge networks, the varying requirements for time-sensitive smart factory services, and the variety of drift types, the task of detecting anomalies in real time and coping with drifts while keeping all possible variations is challenging. The objective of this work is to determine the real-time anomaly status of an observation ${\mathbf{r}}$ at time slot $t$, taking into account the temporal changes in the feature distribution $F_{t}(\mathbf{x})$ over time. Our goal is to dynamically update the prediction rule by considering the observed changes in the predictor features $\mathbf{x}$. Table~\ref{notation} provides an overview of the input parameters and variables utilized in our problem.

\section{Scalable and Adaptable Prediction-Driven Anomaly Detection (SAPDAD)}
\label{solution}
To tackle the problem described in Section~\ref{problem}, we propose an approach to effectively cope with the concept drift of data streams and accomplish real-time and accurate anomaly detection. Fig.~\ref{fig1} shows the overview of our proposed SAPDAD method. \textcolor{black}{This method is designed to handle the high velocity and volume of data streams, two key characteristics of dynamic, large-scale IIoT environments. Our proposed SAPDAD method integrates a multi-source prediction model with an adaptive mechanism for handling concept drift. This ensures sustained detection accuracy for ever-changing data patterns over time. To adapt to new patterns and anomalies in streaming data, we employ an LSTM network for robust time-series prediction and utilize a GA to dynamically optimize the prediction model parameters.} As shown in Fig.~\ref{fig1} SAPDAD comprises two main phases: (i) offline phase and (ii) real-time phase. During the offline phase, historical data is collected, preprocessed, and used to train a predictive model. In the real-time phase, anomalies are detected in streaming data in real time, and the model is retrained once a concept drift occurs. In the following, we provide a more detailed explanation of each phase.

\subsection{Offline Phase}
The offline phase comprises two main components of (i)~Data Preprocessing (ii)~Training and Prediction.

\subsubsection{Data Preprocessing}
\label{preprocess}

The Data Preprocessing module converts the time-series data into the proper form required to be used in a neural network. First, the data is cleansed from errors (e.g., missing values, outliers) and normalized. 
\textcolor{black}{We then perform a correlation analysis to understand the interdependencies among the multi-dimensional features of our data. Given $R^\prime=\left\{\mathbf{r}^{\prime}_{0}, \mathbf{r}^{\prime}_{1}, ..., \mathbf{r}^{\prime}_{M}\right\}$, $\left(\mathbf{r}^{\prime}_{i} \in \mathbb{R}, i \in[0, M]\right)$,
where $\textbf{r}^{\prime}_i$ is a column vector of $D$ features representing a record, the Pearson correlation coefficient $\rho_{r_i^{\prime(l)}, r_i^{\prime(m)}}$ of $r_i^{\prime(l)}$ and $r_i^{\prime(m)}$ across all records is defined as:
\begin{equation}
\rho_{r_i^{\prime(l)}, r_i^{\prime(m)}}=\frac{\operatorname{cov}\left(r_i^{\prime(l)}, r_i^{\prime(m)}\right)}{\sigma_{r_i^{\prime(l)}} \sigma_{r_i^{\prime(m)}}},
\end{equation}
where $\operatorname{cov}\left(r_i^{\prime(l)}, r_i^{\prime(m)}\right)$ is the covariance between two features across all records, and $\sigma_{r_i^{\prime(l)}}$ and $\sigma_{r_i^{\prime(m)}}$ are their standard deviations. This analysis is critical in identifying inter-feature dependencies and potential redundancy within the records.
}Second, we employ the PCA method as our preprocessing module to take into account the correlation between the features \cite{abdi2010principal}. 
\textcolor{black}{The PCA process involves the following steps: 
\newline
\textbf{Standardization}: We first standardize $\mathbf{r}_{i}^{\prime}$ to have a mean of 0 and a standard deviation of 1; the standardized $\mathbf{r}_{i}$ is given by:
\begin{equation}
    \mathbf{r}_{i}=\frac{\mathbf{r}_{i}^{\prime}-\mu_i}{\sigma},
\end{equation}
where $\mu_i$ is the mean of $\mathbf{r}_{i}^{\prime}$, and $\sigma$ is its standard deviation.} 
\newline
\textbf{\textcolor{black}{Covariance Matrix Computation:}}
Using PCA, correlated features can be transformed into uncorrelated features called orthogonal features, also known as principal components. Principal components capture the orientations of the data, explaining the highest proportion of variance. The correlations between relevant features are considered during the process of reducing the dimension. We note that the incorporation of dimension reduction can further enhance the efficiency of our proposed method. By leveraging the covariance matrix of data with a high number of dimensions, PCA allows for its projection onto a new space. In this transformed space, the axes are aligned with the eigenvectors of the covariance matrix, prioritized based on the magnitude of their corresponding eigenvalues. This process effectively reduces the data size by retaining only the directions that capture the most informative aspects, i.e., those associated with higher eigenvalues. Given $R=\left\{\mathbf{r}_{0}, \mathbf{r}_{1}, ..., \mathbf{r}_{M}\right\}$, $\left(\mathbf{r}_{i} \in \mathbb{R}, i \in[0, M]\right)$,
where $r_i$ is a column vector of $D$ features representing a record, its normalized covariance matrix is
\begin{equation}
   \Sigma=\frac{1}{M-1} \sum_{i=0}^{M-1}\left(r_i-\tilde{r}\right)\left(r_i-\tilde{r}\right)^T,
\end{equation}
where $\Sigma$ is the $D \times D$ covariance matrix and the average value of $r_i$ is $\tilde{r}=\frac{1}{M} \sum_{i=1}^{M-1} r_{i}$. Its diagonal holds the variance of each individual record, while the off-diagonal values represent the covariances between different record combinations.
\newline
\textbf{\textcolor{black}{Eigenvalue Decomposition:}} By Utilizing eigenvalue decomposition, we can express the covariance matrix $\Sigma$ as
\begin{equation}
\Sigma=V \Lambda V^{-1}.
\end{equation}
Here, the columns of matrix $V$ represent the eigenvectors, while the principal diagonal of matrix $\Lambda$ contains the corresponding eigenvalues of matrix $\Sigma$. Notably, it can be demonstrated that $V^k \in \mathbb{R}^k$ serves as a basis for the subspace of dimensions $\mathbb{R}^k$, which preserves the strongest resemblance to the original subspace.
\newline
\textcolor{black}{\textbf{Component Selection:} We select the top $B$ components that cumulatively explain a significant portion of the variance. This is determined based on the eigenvalues, with larger values indicating a more explained variance.}
\newline
\textcolor{black}{\textbf{Transformation:} Finally, the original data is transformed into a new set of uncorrelated features (principal components) by projecting it onto the eigenvectors.}

Next, the time series of each sample feature is decomposed via time-series decomposition. For feature decomposition, we apply the so-called Time Series Analysis (TSA) decomposition~\cite{seabold2010statsmodels} method, which is a procedure for predicting time-series data based on an additive model. More specifically, the obtained seasonal features of the record are concatenated with the selected features of the record by PCA as an additional input before being fed to the Training and Prediction module, to be explained next.

\subsubsection{Training and Prediction}

The Training module is responsible for training an LSTM model, which can be used in conjunction with the feature decomposition process explained in Section~\ref{preprocess} to make multi-source predictions using a sequence of records. LSTM provides an effective method for analyzing time correlation features of time-series data. Capturing time correlation is critical as anomalies may be related to previous period data. As a result, by analyzing the time correlation present in time-series data, we can effectively detect anomalies. We note, however, that LSTM models inherently fall short of learning complicated seasonal patterns across a given multi-seasonal time-series data. To cope with this, the proposed method explicitly considers the seasonal aspects of the data as inputs to the LSTM model by feature decomposition. 
Extracting and using the seasonal features not only can help the LSTM model learn complicated seasonal patterns but also lead to more efficient and shorter training.
we employ a hyperparameter optimization technique based on GA \cite{yang2020hyperparameter}. This technique involves searching the hyperparameter space using GA to identify the combination of hyperparameters that best fit the data and produce accurate predictions. By leveraging GA-based hyperparameter optimization, we aim to improve the overall performance of our LSTM model. GA identifies optimal hyperparameter values by leveraging information sharing and cooperation among individuals in a population. We employ GA to identify the optimal hyperparameters and LSTM network architecture, including the number of hidden units, training times, gradient threshold, and learning rate, for achieving the best performance \cite{yang2020hyperparameter}.

\begin{figure*}[t]
\centerline{\includegraphics[width=1.75\columnwidth,keepaspectratio]{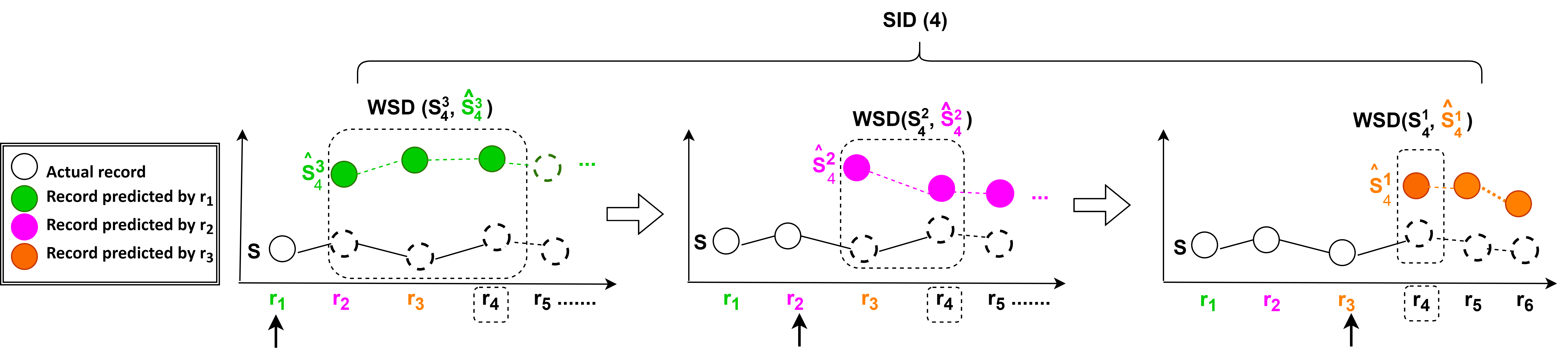}}
\caption{An example demonstrating the calculation of SID
with the maximum prediction horizon length L=3}
\label{figdetail}
\end{figure*}

\begin{algorithm}[t]
\small
\caption{PDAD-SID}\label{alg1}
\begin{algorithmic}[1] 
\REQUIRE Target record $r_{i}$, prediction horizon  $L$, reference series length $S$, records $r_{i-K}$ to $r_{i-1}$, $\text{AP}(i-k)$ for $k$ $\in$ [$1$, $L$]
\ENSURE $\text{SID}(i)$
\STATE $C \leftarrow $1$ ,  \mu \leftarrow $0.5$ $
\FOR{$i=0:s-1$}
\STATE $\text {AP}(i) \leftarrow $0$  $
\ENDFOR
\FOR{$k=1:L$}

\STATE Predict $\mathbf{\hat S}_{i}^k$
\STATE Calculate $\text{WSD}\left(\mathbf{S}^{k}_{i}, \mathbf{\hat{S}}^{k}_{i}\right)$ using Eq.~(\ref{eq2})
\ENDFOR
\WHILE{convergence condition is not met}
\FOR{$i=L:s-1$}
\STATE Calculate $\text{SID}(i)$ using Eq.~(\ref{eq3})
\STATE Calculate $\text{AP}(i)$ using Eq.~(\ref{eq4})
\ENDFOR
\STATE $\mu \leftarrow $ Mean of the SIDs
\STATE $\sigma \leftarrow $ Variance of the SIDs
\STATE $C \leftarrow $1$ $/$ \sigma^2 $
\ENDWHILE
\STATE Update $\text{SID}(i)$ using Eq.~(\ref{eq5})
\end{algorithmic}
\end{algorithm}
\subsection{Real-Time Phase}
\label{realtimephase}
In the real-time phase, the proposed method processes online streams of data generated continuously over time. This phase starts with the utilization of the LSTM model generated during the offline phase to predict the future records of the data stream to compute their anomaly probability utilizing PDAD-SID. PDAD-SID, a multi-source prediction approach, uses a record-to-sequence predictor. This differs from traditional record-to-record predictors~\cite{malhotra2015} by considering a sequence of records for making predictions at different time points rather than only one. Multiple predictions from multiple records form a unified estimated value of the target record, which can be used to determine whether or not it is anomalous, depending on the target record's deviation from the unified estimated value. Algorithm 1 describes the pseudo-code of the proposed PDAD-SID anomaly detection method.
Furthermore, it is crucial to update the anomaly detection model frequently to keep up with the changes in data distribution. Therefore, upon detecting concept drift in the new data streams, the LSTM model will undergo retraining to align with the new IIoT data. As a result, the proposed method can effectively adjust to constantly evolving traffic data patterns and consistently achieve accurate anomaly detection. 
\textcolor{black}{The Real-Time phase comprises two main components of (i)~Anomaly Detection}, and (ii)~Drift Detection and Model Update.

\subsubsection{Anomaly Detection}
The real-time anomaly detection module relies on the calculation of two key metrics, namely, $\text{SID}$ metric and Anomaly Probability (AP), which are explained below. 

\paragraph{SID Metric}
Let $\mathbf {r}_{i}$ and  $\mathbf{r}_{j}$ denote records $i$ and $j$, respectively, $D$ be the dimension of each record, and $r^{(l)}_{i}$ be the component $l$ of record $i$. We obtained the Record Distance (RD) between records $i$ and $j$ as follows:
\begin{equation}
\text{RD}(\mathbf{r}_{i},\mathbf{r}_{j})=\frac{1}{D} \sum_{l=1}^{D}(r_{i}^{(l)}-r_{j}^{(l)})^2.
\label{eq1}
\end{equation}
Next, let us define sequence $\mathbf{S}^{N}_{i}$, $\forall i \ge N$, of length $N$, which comprises $N$ consecutive records ending at record $i$:
\begin{equation}
\mathbf{S}^{N}_{i}=(\mathbf{r}_{i-N+1},...,\mathbf{r}_{i-1},\mathbf{r}_i).
\label{eq11}
\end{equation}
In order to obtain a uniform scale for measuring the distance between two sequences, the sequence distance metric should be independent of the sequence length. Thus, we calculate the distance between sequence $\mathbf{S}^{N}_{i}$ of actual record values and sequence $\mathbf{\hat{S}}^{N}_{i}$ of predicted record values via Weighted Sequence Distance (WSD), which is defined as follows: 
\begin{equation}
\begin{aligned}
&\text{WSD}\left(\mathbf{S}^{N}_{i}, \mathbf{\hat{S}}^{N}_{i}\right)=\frac{\sum_{m=1}^{N} e^{(N-m)} \cdot \text{RD}\left(\mathbf{r}_{i-m+1}, \mathbf{\hat{r}}_{i-m+1}\right)}{\sum_{m=1}^{N}e^{(N-m)}},
\end{aligned}
\label{eq2}
\end{equation}
where $\mathbf{r}_{i-m+1}$ and $\mathbf{\hat{r}}_{i-m+1}$ are the actual and predicted values of record $i-m+1$. In the definition of WSD function, time decay weight $e^{(N-m)}$ is associated with record $i-m+1$ in order to assign a greater level of reference to more recent records. 

To quantify the deviation of the target record $i$ from the estimated value, we define a new metric called SID, which is given by: 
\begin{equation}
\begin{aligned}
&\text{SID}\left(i\right)=\frac{ \sum_{k=1}^{L}    P(i-k) \cdot \text{WSD}(\mathbf{S}_i^{k},\mathbf{\hat{S}}_i^{k})}{\sum_{k=1}^{L}P(i-k)},
\end{aligned}\label{eq3}
\end{equation}
where $L$ is the prediction horizon, which determines the maximum length of the predicted sequence, and $P(i-k)$ denotes the probability that record $i-k$ is normal. The proposed SID metric alleviates the impact of anomalous records within the historical data by weighting the predictions (see Eq.~(\ref{eq3})). Fig.~\ref{figdetail} provides a visual representation of the calculation of SID in a specific case where
$L = 3$. For example, let us assume that we aim to calculate the SID of target 4 for $L = 3$. Once the actual record $\mathbf {r}_{1}$ arrives, the model predicts the next three records $\mathbf{r}_{2}$, $\mathbf{r}_{3}$, and $\mathbf{r}_{4}$. Similarly, once record $\mathbf{r}_{2}$ arrives, the model predicts $\mathbf{r}_{3}$, $\mathbf{r}_{4}$, and $\mathbf{r}_{5}$. Likewise, once $\mathbf{r}_{3}$ arrives, the model predicts $\mathbf{r}_{4}$, $\mathbf{r}_{5}$, and $\mathbf{r}_{6}$. A three-step process is carried out to calculate $\text{SID}(4)$ of the target $4$. First, we calculate the WSD of the actual and matching predicted sequences up to target $4$, i.e., the distances between $\mathbf{S}_{4}^1$ and $\mathbf{\hat S}_{4}^1$, between $\mathbf{S}_{4}^2$ and $\mathbf{\hat S}_{4}^2$, and between $\mathbf{S}_{4}^3$ and $\mathbf{\hat S}_{4}^3$ and then normalize it by $\sum_{m=1}^{N}e^{(N-m)}$, as shown in Eq.~(\ref{eq2}). Subsequently, SID is obtained as the normalized weighted sum of WSDs, see Eq.~(\ref{eq3}).

\paragraph{Anomaly Probability}

Given that the range of the SID metric is application-specific, we aim to map the value of $\text{SID}(i)$ of target $i$ to its probability $\text{AP}(i)$ of being anomalous as follows: 
\begin{equation}
\text{AP}(i)=\varphi(\frac{1}{1+e^{-C\left(\text{SID}(i)-\mu\right)}}),
\label{eq4}
\end{equation}
where $\varphi(\cdot)$ is the logistic mapping function, $\mu$ is the mean of the SIDs of some of the targets, $C$ is the logistic growth rate, which is set to $1/{\sigma^{2}}$, where $\sigma$ is the variance of the SIDs of some of the targets. To determine the optimal $C$ and $\mu$, we use an automated iterative algorithm, which is applied to the training data. As time passes and the probability of the targets becomes more accurate, we calculate $\mu$ and $C$ according to the mean and standard deviation of previous SIDs, respectively, until the algorithm converges to stable values of $\mu$ and $C$.
Using Eq.~\eqref{eq4} to map SIDs to APs not only allows for a better distinguishing between normal and anomalous data, but it helps the values of APs become less application-specific.  
Finally, in order to assign larger weights for normal records in the calculation of SID, we substitute $P(i-k)$ in Eq.~(\ref{eq3}) with $(1- \text{AP}(i-k))$. Thus, Eq.~\eqref{eq3} can be re-written as follows:
\begin{equation}
\begin{aligned}
&\text{SID}\left(i\right)=\frac{ \sum_{k=1}^{L}    (1-\text{AP}(i-k)) \cdot \text{WSD}(\mathbf{S}_i^{k},\mathbf{\hat{S}}_i^{k})}{\sum_{k=1}^{L}(1-\text{AP}(i-k))}.
\end{aligned}
\label{eq5}
\end{equation}

\textcolor{black}{ 
Algorithm 1 utilizes prediction horizon $L$, reference series length $S$, and historical records $\left(r_{i-K}\right.$ to $\left.r_{i-1}\right)$ to calculate SID for real-time anomaly detection. Initially, logistic growth rate $C$ and mean $\mu$ are set, and anomaly probabilities AP are initialized (lines 1 to 4 of Algorithm 1). For each incoming record $\mathbf{r}_{i}$ in the data stream, the algorithm utilizes the LSTM model generated during the offline phase to predict future records. This involves predicting multiple steps ahead, up to the prediction horizon $L$ (lines 5 to 6 of Algorithm 1).
For each predicted step ($k$ from 1 to $L$), the algorithm calculates WSD using Eq.~(\ref{eq2}) (lines 7 in Algorithm 1). 
The SID is then updated using Eq.~(\ref{eq3}), which involves the weighted sum of WSDs. This step provides a measure of the deviation of the target record from the estimated value, considering the historical context and the probability that previous records were normal.
The SID values are mapped to AP using the logistic mapping function described in Eq.~(\ref{eq4}). It refines SID and AP iteratively until convergence and dynamically adjusts the parameters (lines 10 to 13 of Algorithm 1). Likewise, the algorithm iteratively updates parameters $\mu$ and $\sigma$ until convergence, allowing it to adapt to changing data patterns (lines 14 to 16 of Algorithm 1). The resulting SID values are then updated to assign larger weights for normal records in the calculation of SID, the probability of a record being normal, denoted as $(1-\text{AP}(i-k))$ is substituted in the calculation of SID (line 18 of Algorithm 1).
}

\begin{algorithm}[!t]
\small

\caption{RealTimeOAW}\label{alg2}
\begin{algorithmic}[1] 
\REQUIRE R: Data stream, LSTM: Pre-trained LSTM model, $L$: Prediction horizon, $S$: Series length, $A_{th}$: Alert threshold, $D_{th}$: Drift threshold, $L_s$: Fixed sliding window size, $L_a$: Max adaptive window size

\ENSURE AR
\STATE $ Condition \leftarrow Normal  $ 
\STATE   $AdaptWin$ $\leftarrow  \emptyset   $
\FOR {$r_{i}$ in R }
\STATE  $PDAD-SID(i,L,s) $
\STATE $W_{i} \leftarrow$  Window with last $L_s$ records 
\STATE $ARWin_{i} \leftarrow $$AR(W_{i})$$   $ \COMMENT{Current window AR} 
\STATE $ARWin_{i-L_s} \leftarrow $$AR(W_{i-L_s})$$   $  \COMMENT{Last window AR}
\IF {($Condition=Normal$) and ($ARWin_{i}$  $\geq$  $A_{th}$ * $ARWin_{i-L_s}$) }
    \STATE  $AdaptWin \leftarrow  AdaptWin    \cup  {r_{i}}$ \COMMENT{collecting new records}
    \STATE $Condition \leftarrow Alert $
\ELSIF {($Condition=Alert$)} 
    \STATE $L'_a\leftarrow  Len(AdaptWin )  $ 
    \IF {$ARWin_{i}$  $\geq$  $D_{th}$ * $ARWin_{i-L_s})$}  
    \STATE $Condition \leftarrow Drift$
    \STATE $j\leftarrow  $i$ $  \COMMENT{Determine the AR for the first concept drifted window}
    \STATE $UpdatedLSTM \leftarrow  $retrain LSTM on $AdaptWin$ $ $  
    \ELSIF{($ARWin_{i}$  $<$  $A_{th}$ * $ARWin_{i-L_s}$ or $L'_a$==$L_a$) \COMMENT{False alarm} }  
    \STATE $AdaptWin\leftarrow $0$ $ 
    \STATE $Condition \leftarrow Normal  $ 
    \ELSE
    \STATE  $AdaptWin  \leftarrow  $AdaptWin $  \cup \{r_i\}$  
\ENDIF
\ELSIF {($Condition=Drift$)} 
    \STATE $L'_a\leftarrow  Len(AdaptWin )  $ 
    \IF{($ARWin_{i}$  $\geq$  $A_{th}$ * $ARWin_{j+L_s}$ or $L'_a$==$L_a$)}
    

    \STATE $ UpdatedLSTM \leftarrow  $retrain UpdatedLSTM on $AdaptWin$ $ $  
    \STATE $   AdaptWin \leftarrow  \emptyset $            
    \STATE $ Condition \leftarrow Normal  $ 
    \ELSE
        \STATE  $AdaptWin  \leftarrow  AdaptWin    \cup  \{r_i\}$ 
    \ENDIF
\ENDIF
\ENDFOR 
\RETURN $AR$
\end{algorithmic}
\end{algorithm}

\begin{algorithm}
\small
\caption{HyperParametersOptimization}\label{alg3}
\begin{algorithmic}[1] 
\REQUIRE Space: configuration space, MaxTime: iteration for hyperparameter search
\ENSURE $Optimized_{HP}$: the detected optimal hyperparameter values\\ MinAR: the average overall AR 
\STATE $MinAR\leftarrow 0$
\FOR{$j=1:MaxTime$}
\STATE $ $$A_{th}$,$D_{th}$,$L_s$,$L_a$$  \leftarrow  Genetic Algorithm(Space)$ 
\STATE   $ AR \leftarrow  RealTimeOAW (R, LSTM,L,s, $$A_{th}$,$D_{th}$,\\$L_s$,$L_a$$  )  $  \COMMENT{ Evaluate the current hyperparameters configuration}
\IF{$MinAR$ $\ge$ $AR$}
    \STATE $MinAR \leftarrow  $AR$ $
    \STATE $Optimized_{HP}\leftarrow  $$A_{th}$,$D_{th}$,$L_s$,$L_a$$ $ 
    \ENDIF   
\ENDFOR
\RETURN $MinAR$, $Optimized_{HP}$

\end{algorithmic}
\end{algorithm}

\subsubsection{Drift Detection and Model Update}

IIoT streaming data may be subject to a variety of data distribution changes, which may occur in dynamic IIoT environments. Thus, the anomaly detection model should be updated frequently in order to keep up with the changes in data distribution. To cope with this issue, we propose a RealTimeOAW method to detect the concept drift and adapt to it. Our proposed RealTimeOAW algorithm combines sliding and adaptive window-based methods with performance-based approaches. The RealTimeOAW algorithm is presented in Algorithm 2, which detects concept drift within streaming data and subsequently incorporates new records to update the model. Algorithm 3 (HyperparameterOptimization Algorithm) uses GA to tune and optimize the hyperparameters of the RealTimeOAW algorithm.

\begin{figure}[!t]
\centerline{\includegraphics[width=9cm,height=5cm,keepaspectratio]{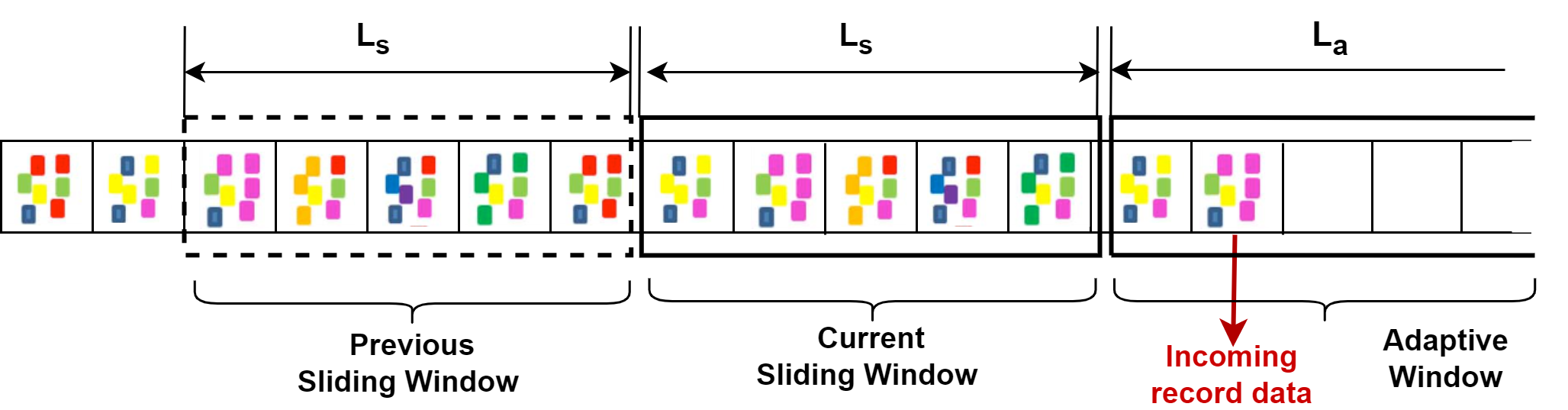}}
\caption{An example of the sliding window and adaptive window of the proposed RealTimeOAW algorithm.}
\label{windowexample}
\end{figure}

Our proposed RealTimeOAW algorithm exploits two types of windows: (i) sliding window for the detection of the concept drift and (ii) adaptive window for storing newly arrived records. The sliding window has a fixed size of $L_s$, which contains the most recent records. 
The adaptive window, which has a maximum size of $L_a$, can adapt its size dynamically based on the number of arriving records. When a new record arrives, it is added to the adaptive window. If the number of records in the adaptive window exceeds $L_a$, the oldest records are removed from the window. By keeping a window of recently arrived records, the adaptive window ensures that the algorithm can adapt to the changes in data distribution.
The sliding window $W_{i}$ associated with data record $r_{i}$ contains $L_s$ records from record $i$ – $L_s$ to record $i$. The Anomaly Rate (AR) within the window can be calculated by considering both the anomaly probability and the number of anomalous data records in the corresponding window. If the anomaly probability of each record data exceeds a pre-defined threshold $T$, i.e., $\text{AP}(i)>T$, then record $i$ is considered as an anomalous record. Let $W$ denote the set of the anomaly probabilities in the given sliding window $W_{i}$ as follows: 
\begin{equation}
\text{W}=\left\{AP(i-L_s),\ldots, AP(i-1),AP(i)\right\}.
\end{equation}
We then define $W_A$ as the set of APs ($\in W$) that are greater than the given threshold $T$: 
\begin{equation}
\begin{aligned}
& W_A=\left\{AP(i) \in W \mid AP(i)> T\right\} .
\end{aligned}
\end{equation}
Then, we calculate AR as follows:
\begin{equation}
\text{AR}=\frac{|W_A|}{|W|}.
\end{equation}
\textcolor{black}{To detect the concept drift, we define thresholds $A_{th}$ and $D_{th}$, which are used to trigger the alert and drift levels, respectively. More specifically, when the difference between the ARs of the current sliding window $i$ and the sliding window $i$ - $L_s$ exceeds $A_{th}$, the alert level is activated, prompting the adaptive window to commence the collection of new incoming data records (lines 5 to 10 of Algorithm 2). Similarly, a drift is detected once the difference between the ARs of the current sliding window $i$ and the sliding window $i$ - $L_s$ exceeds $D_{th}$. At this point, the old learner is retrained on the newly collected records in the adaptive window (lines 12 to 16 of Algorithm 2)}. Specifically, the system enters the retraining phase when one of the subsequent criteria is met:
If the current state is normal, the system enters the alert state when the AR of the current sliding window exceeds $A_{th} \times ARWin_{i-L_s}$.
If the current state is alert, the system enters the drift state when the AR of the current sliding window exceeds $D_{th} \times ARWin_{i-L_s}$.
If the AR of the current window is in either the alert state or the drift state and it does not exceed its threshold or the maximum adaptive window size $L_a$ is reached, the system switches back to the normal state and releases the adaptive window. Once the adaptive window is filled with enough new records, the algorithm retrains an LSTM model on the combined data from the adaptive window and the fixed sliding window. After retraining, the system switches back to the normal state, and the adaptive window is released.
A visual representation of the sliding and adaptive windows' operations is shown in Fig.~\ref{windowexample}.

To ensure accurate and consistent learning, the adaptive window is designed to continue collecting records until either of the following conditions is met: (1) the new AR exceeds the alert threshold $A_{th}$ in relation to the starting point of the drift, signifying the current learner's inability to handle the new data record and requiring an update, and (2) the adaptive window reaches its maximum capacity of $L_a$, guaranteeing that real-time constraints are met. \textcolor{black}{The learner is then updated with the most recent records within the adaptive window, resulting in increased resilience and restoring the system to its normal state (lines 23 to 32 of Algorithm 2). Conversely, if the sliding window AR ceases to rise or even drops to the normal level during the alert condition, it is considered as a false alarm. Upon the release of the adaptive window, the system reverts to its normal state. This allows for the monitoring of new drift occurrences (lines 18 to 22 of Algorithm 2)}.

In order to achieve optimal performance, the hyperparameters of the LSTM models and RealTimeOAW must be tuned and optimized. Two categories of hyperparameters require tuning in the LSTM models: model design and model training hyperparameters. Model design hyperparameters are set during the model design process and include parameters such as the number of layers, learning rate, and dropout rate. In contrast, model training hyperparameters balance training speed and model performance and include parameters such as batch size and epoch number. It is important to note that these hyperparameters have a direct impact on the structure, effectiveness, and efficiency of the LSTM models.
We note that four parameters play an essential role in the RealTimeOAW algorithm, $A_{th}$, $D_{th}$, $L_s$, and $L_a$. The performance of the RealTimeOAW algorithm is directly influenced by these parameters. In RealTimeOAW, GA is employed to tune these hyperparameters, leading to the creation of an optimal adaptive learner capable of handling both continuous and discrete parameters. The Hyperparameter Optimized Algorithm is shown in Algorithm 3, which uses GA  to identify the optimal combination of hyperparameters that return the smallest overall AR. The detected optimal hyperparameters are subsequently fed to the RealTimeOAW algorithm to build an optimized model to detect anomalies accurately.

\subsection{Time Complexity}

In the following, we present the time complexity of our proposed method. In our proposed method, matrix calculations can be used to formulate the calculation of SID, allowing the calculations to be performed in parallel. As a result, the calculation of WSD and RD can be further accelerated. By leveraging the Scalable Universal Matrix Multiplication Algorithm (SUMMA) \cite{van1997}, we can significantly reduce the time complexity of computing WSD in Eq.~(\ref{eq2}) from O($L^3$) to O($L$). Additionally, the computation of WSD and SID in Eq.~(\ref{eq3}) can be efficiently performed in parallel since both involve a vector-matrix multiplication. This parallelization requires only $L$ arithmetic operations, resulting in a time complexity of O($L^2/L$) = O($L$). Moreover, multiplying the matrices in Eq.~(\ref{eq3}) can be accomplished with a complexity of O($L$). It is important to note that the matrix RD contains $L\times L$ entries of record distance, and each entry is calculated using Eq.~(\ref{eq1}). Therefore, updating RD (upon arrival of a new record) runs with the complexity of O($L^2D/L^2$) = O($D$), where D is the record dimension. Consequently, by incorporating parallelization techniques, the time complexity of computing SID$(i)$ is reduced to O($D + L$). As a result, by increasing the number of dimensions, the time complexity of calculating the SID metric grows linearly.
The PDAD-SID algorithm time calculates RD between each pair of records with a time complexity of O($D$) without any parallelization. Thus, the time complexity for obtaining WSD between two sequences is O($LD$).
Therefore, the time complexity of detection at a single record is O($L^2$), because $L$ sources of predictions are used (see Eq.~(\ref{eq5})). With proper parallelization, the time complexity can be reduced to O($D + L$). Moreover, the training complexity of RealTimeOAW is O($MH$). Here, $M$ denotes the number of records, while $H$ represents the maximum number of iterations for hyperparameter search using the GA.

\section{Performance Evaluation}
\label{evaluation}

In this section, we conduct a performance evaluation of our proposed SAPDAD method. After describing our simulation environment, datasets, and evaluation metrics, we present our findings. Finally, we assess the impact of the anomaly probability threshold $T$ on the performance.

\subsection{Experiment Environment}
\label{environment}

Our evaluations were conducted on a system equipped with a quad-core CPU (model: Intel Core i7-7700, 3.60~GHz) and 16~GB of RAM. The inference module of our prediction model was implemented using the Tensorflow-addons platform (V.: 0.14), while the decomposition module utilized the TSA Decomposition library (V.: 0.4). We conducted a hold-out validation approach, where the initial model training utilized the first 10\% of the data, while the remaining 90\% was reserved for online testing purposes.
The LSTM network structure used for each dataset is based on $[\#Raw Features + \#Seasonal Features, Time Step, \#Raw Features \times L]$. 
 Three different time steps were evaluated, where the seasonality length was 24 (daily seasonality length), 72, or 168 (weekly seasonality length). 
 Dropout was disabled. In order to avoid over-fitting, a weight decay of $6 \times 10^{-6}$ was applied during the training.

\begin{table}[!t]
\caption{\textcolor{black}{Hyperparameter configuration of LSTM and RealTimeOAW on (A) KDDCup99, (B) IoTID20, and (C) WUSTL-IIoT with Optimal Values.}}
\begin{tabular}{|c|l|c|c|c|c|}
\hline
\label{config}
Model & Hyperparameter & Search Range & A & B & \textcolor{black}{C} \\ \hline
LSTM & \begin{tabular}[c]{@{}l@{}}Number\\of\\ Epochs\end{tabular} & [20, 100] & 100 & 87 & \textcolor{black}{100}\\ \cline{2-6} 
& Learning Rate & [0.0001, 0.01] & 0.001 & 0.0001 & \textcolor{black}{0.001}\\ \cline{2-6} 
& Optimizer & [adam, sgd] & adam & adam & \textcolor{black}{adam}\\ \cline{2-6} 
& \begin{tabular}[c]{@{}l@{}}Activation\\Function\end{tabular} & [relu, tanh] & relu & tanh & \textcolor{black}{tanh} \\\cline{2-6} 
& Loss Function & [MSE,MAE] & MSE & MSE & \textcolor{black}{MSE} \\ \cline{2-6} 
& Batch Size & [10, 50] & 32 & 45 & \textcolor{black}{39}\\ \hline
\multicolumn{1}{|l|}{\begin{tabular}[c]{@{}l@{}}RealTime\\OAW\end{tabular}} & $A_{th}$ & (0, 0.1) & 0.092 & 0.089 & \textcolor{black}{0.095} \\ \cline{2-6} 
\multicolumn{1}{|l|}{} & $D_{th}$ & (0, 0.08) & 0.03 & 0.045 & \textcolor{black}{0.034} \\ \cline{2-6} 
\multicolumn{1}{|l|}{} & $L_s$ & {[}100, 600{]} & 270 & 150 & \textcolor{black}{325} \\ \cline{2-6} 
\multicolumn{1}{|l|}{} & $L_a$ & {[}400, 4000{]} & 2500 & 450 & \textcolor{black}{3240} \\ \hline
\end{tabular}
\end{table}

Table~\ref{config} presents the hyperparameters of LSTMs and RealTimeOAWs, which were automatically tuned using GA. The experimental setup consisted of a population size of 70, a crossover rate of 0.7, and a mutation rate of 0.15. Further, we determined the optimal threshold for each of the time-series datasets considering their common properties and validation data. We consider a set of threshold values for anomaly probability, including $T= \left\{0.55, 0.60, 0.65, 0.70, 0.75\right\} $. Our simulation results indicate that the optimal threshold value falls within the range of 0.55 to 0.75, because the large thresholds result in more false anomalies, whereas the small thresholds prolong the execution time. The values of prediction horizon $L$ are selected from $\left\{10,15,20,30\right\}$~\cite{Raeiszadeh2023}.
\begin{table*}
\centering
\caption{\textcolor{black}{Performance comparison of different anomaly detection methods with drift adaptation.}}
\label{compare}
\begin{tabular}{|l|c|c|c|c|c|c|c|c|c|} 
\hline
{Model} & \multicolumn{3}{c|}{KDDCup99} & \multicolumn{3}{c|}{IoTID20} & \multicolumn{3}{c|}{\textcolor{black}{WUSTL-IIoT}} \\ 
\cline{2-10}
& \multicolumn{1}{l|}{AUC (\%)} & \multicolumn{1}{l|}{\begin{tabular}[c]{@{}l@{}}Avg. Exec.\\ Time (ms)\end{tabular}} & \multicolumn{1}{l|}{\begin{tabular}[c]{@{}l@{}}Proc. Rate\\ (Rec/Sec)\end{tabular}} & \multicolumn{1}{l|}{AUC (\%)} & \multicolumn{1}{l|}{\begin{tabular}[c]{@{}l@{}}Avg. Exec.\\ Time (ms)\end{tabular}} & \multicolumn{1}{l|}{\begin{tabular}[c]{@{}l@{}}Proc. Rate\\ (Rec/Sec)\end{tabular}} & \multicolumn{1}{l|}{\textcolor{black}{AUC (\%)}} & \textcolor{black}{\makecell{Avg. Exec.\\ Time (ms)}}& \textcolor{black}{\makecell{Proc. Rate\\ (Rec/Sec)}} \\ 
\hline
ORUNADA \cite{Dromard2017} & 78.83 & 2.3491 & 427 & 76.72 & 1.2369 & 917 & \textcolor{black}{77.19} & \textcolor{black}{1.9861} & \textcolor{black}{501} \\ 
\hline
\textcolor{black}{A-Detection\cite{wang2021concept} }& \textcolor{black}{80.13} & \textcolor{black}{0.4755} & \textcolor{black}{1896} & \textcolor{black}{79.07} & \textcolor{black}{0.4580} & \textcolor{black}{1970} & \textcolor{black}{78.58} & \textcolor{black}{0.4572} & \textcolor{black}{1973} \\ 
\hline
ASTREAM \cite{yang2022astream} & 88.11 & 1.8870 & 531 & 78.91 & 1.0670 & 1117 & \textcolor{black}{78.68} & \textcolor{black}{2.2210} & \textcolor{black}{450} \\ 
\hline
Online RNN-AD \cite{saurav2018online} & 77.73 & 3.2433 & 308 & 74.03 & 3.0122 & 332 & \textcolor{black}{76.28} & \textcolor{black}{4.2452} & \textcolor{black}{235} \\ 
\hline
PDAD-SID \cite{Raeiszadeh2023} & 82.15 & 0.0305 & 2011 & 79.28 & 0.0119 & 2118 & \textcolor{black}{80.31} & \textcolor{black}{0.0652} & \textcolor{black}{1995} \\ 
\hline
Proposed SAPDAD & 89.71 & {0.0525} & {1945} & {80.66} & {0.0299} & {2021} & \textcolor{black}{83.08} & \textcolor{black}{0.0801} & \textcolor{black}{1984} \\
\hline
\end{tabular}
\end{table*}

\subsection{Description of Datasets}
\label{dataset}

Our evaluations are carried out on three high-dimensional anomaly detection datasets, namely, the KDDCup99, IoTID20, and WUSTL-IIoT.
\begin{itemize}
  \item KDDCup99 \cite{spinosa2008cluster} is a balanced intrusion detection dataset, which includes a variety of hand-injected anomalies (i.e., network attacks) into the normal network data. The dataset has 43 dimensions, with an anomaly rate of 1.77\%. 
\item IoTID20 \cite{ullah2020scheme} is an IoT traffic dataset for anomaly detection where the distribution of records is unbalanced, with 94\% categorized as normal and the remaining 6\% labeled as anomalous. The dataset was generated by employing virtual machines representing both normal network behavior and attack scenarios, mimicking IoT services through the utilization of the node-red tool. Subsequently, features were extracted using the Information Security Center of Excellence (ISCX) flow meter program. The reduced IoTID20 dataset used in this work contains 6253 records, which were randomly sampled based on the time slot at a rate of one data record per 10 time slots.

\item  
\textcolor{black}{WUSTL-IIoT~\cite{zolanvari2019machine} is a network data of IIoT and it is collected from real-world industrial systems. This dataset consists of a variety of IIoT components such as sensors, actuators, Human-Machine Interfaces (HMI), Programmable Logic Controllers (PLC), data loggers, and alarm systems, all aimed at mimicking actual industrial operations.
The dataset is characterized by 41 distinct features, chosen for their variability during different attack phases. The types of cyber attacks represented in this dataset include command injection, Denial of Service (DoS), reconnaissance, and the use of backdoors.}
\end{itemize}

\begin{figure}[htbp]
\begin{center}
\subfigure[]
{\includegraphics[width=0.75\columnwidth,keepaspectratio]{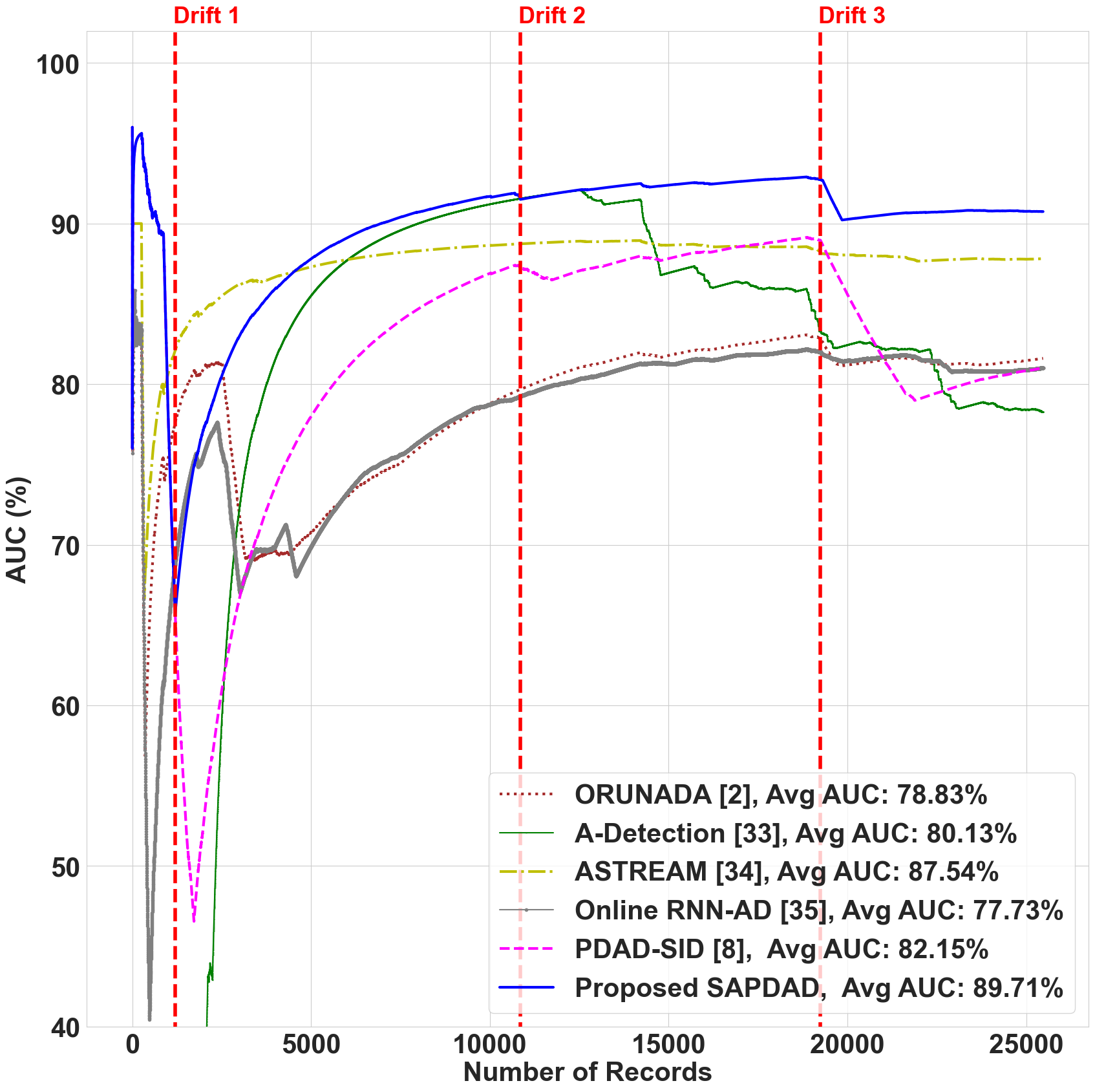}
\label{AUC-a}
}
\subfigure[] 
{\includegraphics[width=0.75\columnwidth,keepaspectratio]{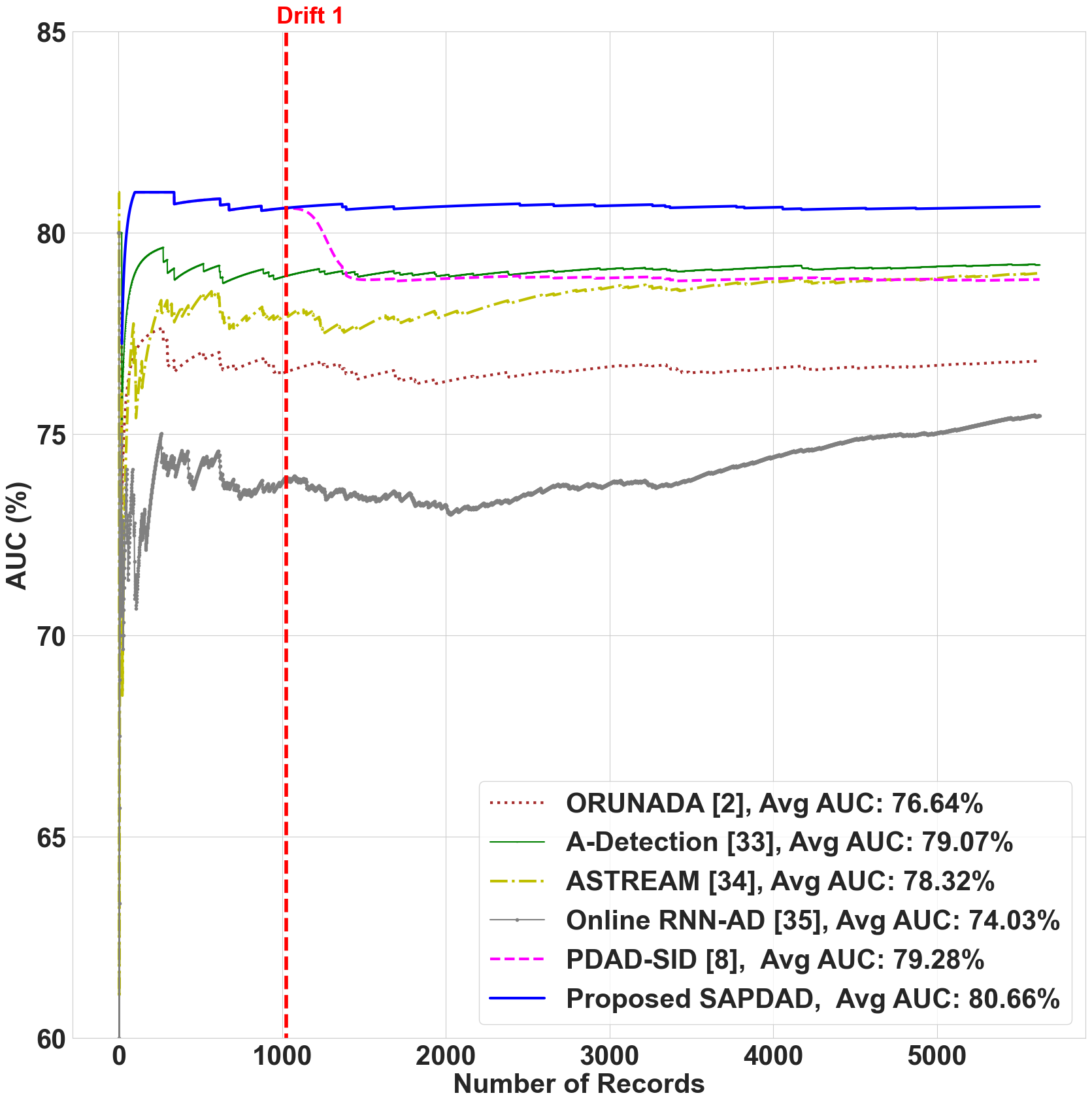}
\label{AUC-b}
}
\subfigure[] 
{\includegraphics[width=0.75\columnwidth,keepaspectratio]{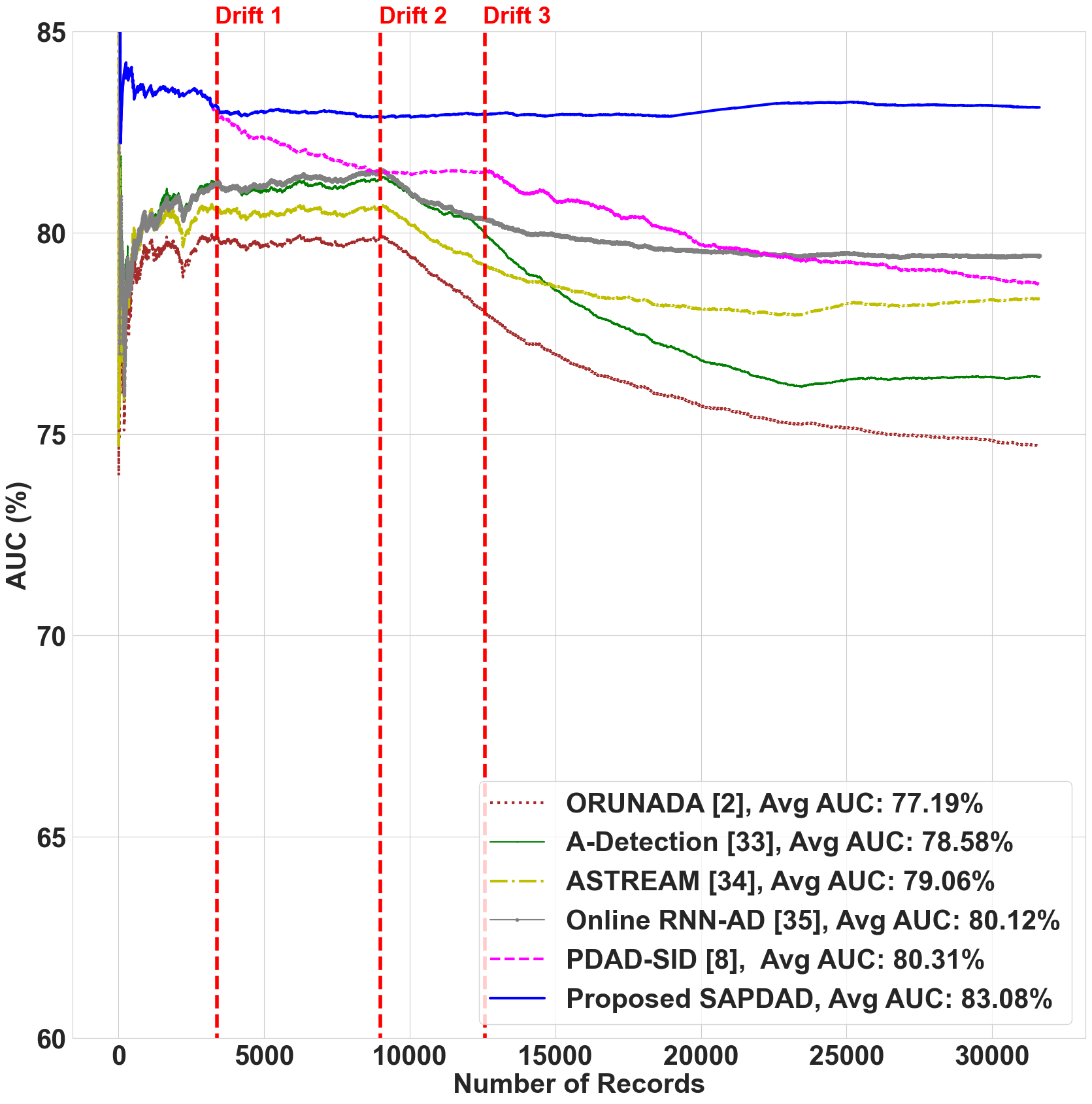}
\label{AUC-c}
}
\caption{\textcolor{black}{AUC of different anomaly detection methods with drift adaptation vs. the number of records for (a) KDDCup99, (b) IoTID20, and (c) WUSTL-IIoT datasets.}}
\label{AUC}
\end{center}
\end{figure}

\subsection{Evaluation Metrics}
\label{metric}

In the following, we present our evaluation metrics. First, we evaluate the effectiveness of different detection methods under study by the so-called Area Under Curve (AUC), which measures the accuracy. A large value of AUC indicates that the model has a better measure of separability, which is key for anomaly detection to meet our accuracy and drift adaptive requirements. Second, we consider two Quality of Service (QoS) parameters related to machine learning and data analytics, namely, average execution time and processing rate. Average execution time is the time required for processing a single data record, including anomaly detection, drift detection, and model updating. A short execution time is crucial to meet the real-time and time-efficiency requirements of the anomaly detection system. Moreover, the processing rate measures the number of processed data records per second. A large processing rate is essential for scaling the proposed method to a large number of data records, i.e., scalability requirement. 
\textcolor{black}{It is worthwhile to note that that in our evaluations, we do not differentiate between anomaly types, because our algorithm is designed to improve anomaly detection efficiency, ensuring its applicability across various scenarios.}

\subsection{Results}

\label{result}
\label{evaluationresult}

We compare the performance of our proposed SAPDAD anomaly detection algorithm with several benchmarks, namely, ORUNADA \cite{Dromard2017}, A-Detection~\cite{wang2021concept}, ASTREAM \cite{yang2022astream}, Online RNN-AD \cite{saurav2018online}, and PDAD-SID~\cite{Raeiszadeh2023}.

Table~\ref{compare} shows a comparison between our proposed SAPDAD anomaly detection algorithm and our benchmarks for three datasets. \textcolor{black}{We observe from the table that the proposed method outperforms all other methods in terms of AUC, achieving a score of 89.71\% on the KDDCup99 dataset, 80.66\% on the IoTID20 dataset, and 83.08\% on the WUSTL-IIoT dataset.} \textcolor{black}{In comparison, ORUNADA \cite{Dromard2017}, A-Detection~\cite{wang2021concept}, ASTREAM \cite{yang2022astream}, Online RNN-AD \cite{saurav2018online} have a significantly lower AUC scores of 78.83\%, 80.13\% 88.11\%, and 77.73\% on the KDDCup99 dataset, 76.72\%, 79.07\%, 78.91\%, and 74.03\% on the IoTID20 dataset, and 77.19\%, 78.58\%, 78.68\%, and 76.28\% on the WUSTL-IIoT dataset, respectively}. \textcolor{black}{The high AUC achieved by the proposed SAPDAD demonstrates its robustness and also reflects its ability to adapt to evolving data patterns. Next, AUC vs. the number of records is shown in Fig.~\ref{AUC}, where we observe that as the data volume increases, the proposed SAPDAD method maintains a consistent AUC, indicating its resilience to concept drift. In contrast, other methods such as Online RNN-AD~\cite{saurav2018online}, while efficient in processing rate, show variations in their AUC performances.}
According to Fig.~\ref{AUC-a}, although a small drift occurred in the KDDCup99 dataset at the early stage of the experiment, all methods were able to adjust to it, but at different rates. Our proposed method adapted to all drifts detected at data records 1201, 10854, and 19245, achieving the highest AUC of 89.71\%, while the PDAD-SID model's AUC dropped to only 82.15\% without drift adaptation.
Similarly, in the IoTID20 dataset, as illustrated in Fig.~\ref{AUC-b}, our proposed method achieved an AUC of 80.66\%, highlighting its ability to adapt to a subtle concept drift detected at data record 1023. In comparison, the PDAD-SID model~\cite{Raeiszadeh2023} had a slightly lower AUC of 79.28\% without drift adaptation. \textcolor{black}{ As illustrated in Fig.~\ref{AUC-c}, our proposed method achieves an AUC of 83.08\%, demonstrating its ability to adapt to a concept drift detected at data records 3380, 9000, and 12584. In comparison, the PDAD-SID method~\cite{Raeiszadeh2023} has a slightly lower AUC of 80.31\% without any drift adaptation.} \textcolor{black}{Therefore, while the PDAD-SID model~\cite{Raeiszadeh2023} shows a comparable efficiency in terms of execution time, it's lower AUC and processing rate suggests a trade-off between speed and accuracy. This trade-off is evident in Fig.~\ref{AUC}, where the performance of PDAD-SID falls below SAPDAD, especially for the IoTID20 and WUSTL-IIoT datasets, which present more complex and unbalanced data characteristics.} 

\textcolor{black}{Table~\ref{compare} also presents the average execution time performance of different algorithms under consideration. The proposed model outperforms ORUNADA \cite{Dromard2017}, A-Detection~\cite{wang2021concept}, ASTREAM \cite{yang2022astream}, and Online RNN-AD \cite{saurav2018online} with significantly shorter average execution times of only 0.0525 ms, 0.0299 ms, and 0.0801 ms, respectively, across the three datasets.} The reason for this is mainly due to PDAD-SID's efficiency and sliding window strategy. In terms of record processing rate, ORUNADA \cite{Dromard2017}, ASTREAM \cite{yang2022astream}, and Online RNN-AD \cite{saurav2018online} outperform the proposed model. However, their AUC falls significantly short compared to the performance achieved by our proposed method. Additionally, in contrast to PDAD-SID, the proposed method demonstrates a significantly improved processing rate of 1945, 2021, and 1984 records per second in the three datasets.
While PDAD-SID exhibits shorter execution times for each record compared to the proposed model, its AUC is notably lower than the proposed method. Hence, based on our findings, setting the anomaly probability threshold $T$ to 0.65, 0.70, and 0.70, and the prediction horizon $L$ to 10, 15, and 15 for KDDCup99, IoTID20, and WUSTL-IIoT, respectively lead to a suitable accuracy-efficiency trade-off for practical scenarios (see Table~\ref{compare}).
\textcolor{black}{
The combination of high AUC scores and the graphical trends in Fig.~\ref{AUC} underline SAPDAD's superior performance in adapting to concept drifts, a critical aspect in dynamic IIoT environments. The proposed SAPDAD not only achieves a high accuracy (as indicated by AUC) but also maintains efficiency in terms of execution time, thus making it highly suitable for real-time applications. The consistent performance of the proposed method across various datasets KDDCup99, IoTID20, and WUSTL-IIoT highlights its robustness and applicability in a wide range of IIoT environments.}

 \textcolor{black}{In addition to high accuracy, the SAPDAD method demonstrates a notable efficiency in processing IIoT data streams. This efficiency is primarily achieved through its unique architectural design and optimization. Specifically, the use of a lightweight LSTM network structure coupled with a strategic hold-out validation approach ensures swift data processing without compromising the ability to learn from complex data patterns. Furthermore, the application of weight decay during training helps prevent overfitting, thereby maintaining agility in processing new data. Moreover, the results demonstrate operational efficiency, which is essential to run real-time anomaly detection in resource-constrained environments. This is evidenced by the obtained execution times shown in Table~\ref{compare}. Additionally, by dynamically adjusting to concept drifts and evolving data patterns, the proposed SAPDAD method reduces the need for frequent retraining and/or manual intervention, thereby saving computational resources and time.}

\begin{figure}[!t]
\begin{center}
\includegraphics[width=0.85\columnwidth,keepaspectratio]{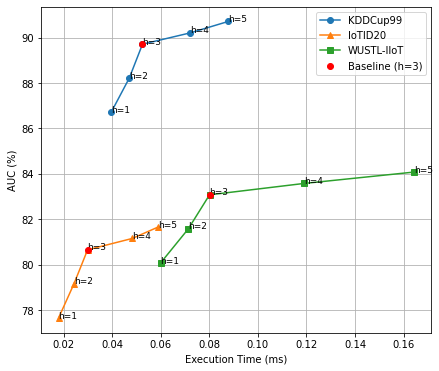}
\caption{\textcolor{black}{AUC vs. execution time for different values of the number $h$ of LSTM hidden layers.}}
\label{layers}
\end{center}
\end{figure}

\begin{figure*}[htbp]
\begin{center}
\subfigure[]
{\includegraphics[width=0.32\textwidth]{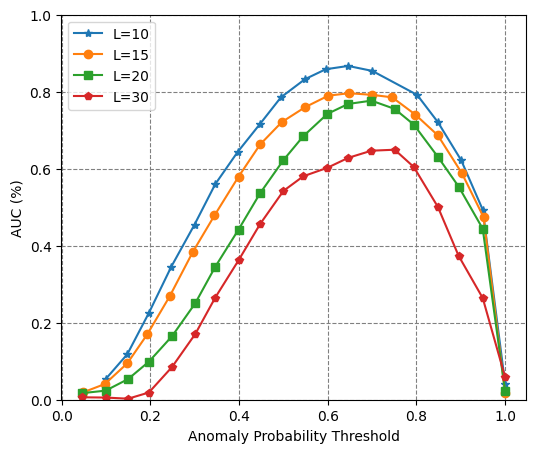}
\label{windowsize-a}
}
\subfigure  []    
{\includegraphics[width=0.32\textwidth]{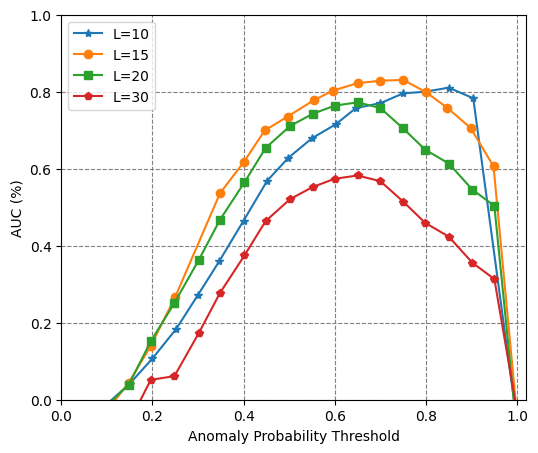}
\label{windowsize-b}
}
\subfigure  []    
{\includegraphics[width=0.32\textwidth]{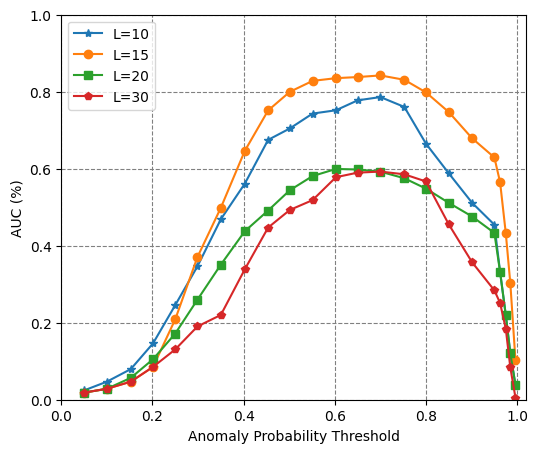}
\label{windowsize-c}
}
\caption{\textcolor{black}{Average AUC vs. anomaly probability threshold $T$ for different values of prediction horizon $L$ in (a) KDDCup99 dataset, (b) IoTID20 dataset, and (c) WUSTL-IIoT dataset.}}
\label{windowsize}
\end{center}
\end{figure*}

\begin{figure*}[htbp]
\begin{center}
\label{TimeLplot}
\subfigure[]
{\includegraphics[width=0.32\textwidth]{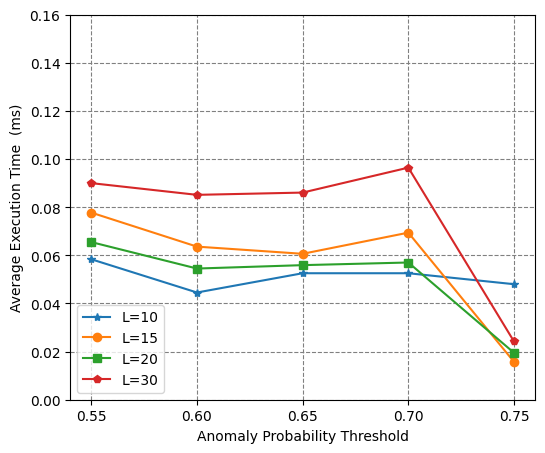}
\label{timeL-a}
}
\subfigure []   
{\includegraphics[width=0.32\textwidth]{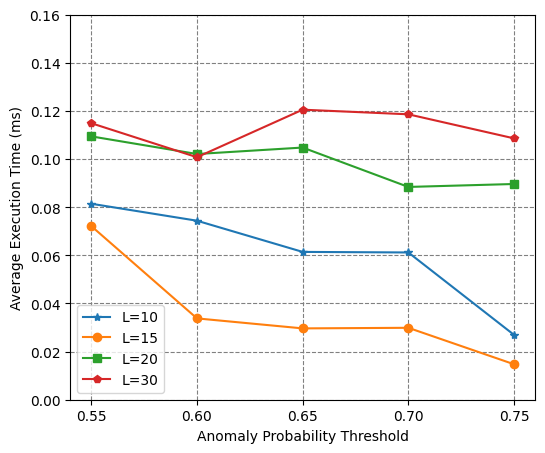}
\label{TimeL-b}
}
\subfigure []
{\includegraphics[width=0.32\textwidth]{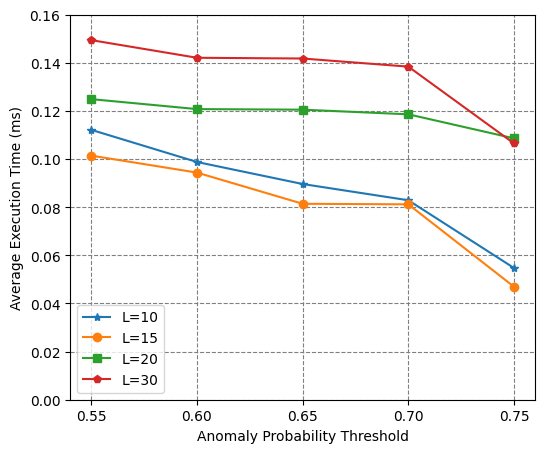}
\label{TimeL-c}
}
\caption{\textcolor{black}{Average execution time vs. anomaly probability threshold $T$  for different values of prediction horizon $L$ in (a) KDDCup99 dataset, (b) IoTID20 dataset, and (c) WUSTL-IIoT dataset.}}
\label{TimeLsize}
\end{center}
\end{figure*}

\textcolor{black}{Next, we assess the correlation between accuracy and execution time. Fig.~\ref{layers} depicts AUC vs. execution time for different values of the number $h$ of LSTM hidden layers for KDDCup99, IoTID20, and WUSTL-IIoT datasets. We observe from the figure that the AUC exhibits a positive correlation with the number $h$ of hidden layers, which was expected. More specifically, as $h$ increases, the obtained AUC also increases, which comes at the expense of a larger execution time. The increased execution time is a direct consequence of incurring complexity to the underlying LSTM model.
Fig.~\ref{layers} is key to help the decision-makers make a trade-off between accuracy and cost.}

Next, we examine the trade-off between AUC and execution time for different values of input parameters of anomaly probability threshold $T$, prediction horizon $L$, and input dimension size $D$. 
\paragraph{AUC}
Figure~\ref{windowsize} illustrates the AUC of the proposed SAPDAD algorithm for different values of anomaly probability threshold $T$ and prediction horizon $L$. Given that our proposed algorithm carries out a probabilistic detection rather than classification, we use AUC as our metric, which offers the advantage of being threshold-independent. The values of anomaly probability threshold $T$ and prediction horizon $L$ was selected from $\left\{0.55, 0.60, 0.65, 0.70, 0.75\right\}$ and $\left\{10, 15, 20, 30\right\}$. As shown in Fig.~\ref{windowsize-a}, the highest AUC was achieved for $L$=10 for a wide range of anomaly probability threshold values. Therefore, we set the prediction horizon $L$ to 10 in the rest of our evaluations. For a fixed prediction horizon $L$, the optimal anomaly probability threshold $T$ was 0.65 for the KDDCup99 dataset because it resulted in improved AUC. Hence, we consider the data records with an anomaly probability of greater than 0.65 as anomalous. For the IoTID20 dataset, as shown in Fig.~\ref{windowsize-b}, $L$=15 leads to the highest AUC with an anomaly probability threshold $T$ of 0.70. \textcolor{black}{For the WUSTL-IIoT dataset, as shown in Fig.~\ref{windowsize-c}, $L$=15 leads to the highest AUC with an anomaly probability threshold $T$ of 0.70}. We note that a higher anomaly probability threshold results in fewer false anomalies, whereas a lower threshold increases the false negatives. \textcolor{black}{Therefore, we determined the anomaly probability threshold $T$ to 0.65, 0.70, and 0.70 in the KDDCup99, IoTID20, and WUSTL-IIoT datasets, respectively.}
\paragraph{Execution Time}
Figure~\ref{TimeLsize} illustrates the execution time for different values of anomaly probability threshold $T$ and prediction horizon $L$. \textcolor{black}{According to Fig.~\ref{timeL-a}, Fig.~\ref{TimeL-b}, and Fig.~\ref{TimeL-c}, it is evident that the average execution time of the proposed SAPDAD decreases as the anomaly probability threshold $T$ increases for a given prediction horizon $L$}. This is due to the fact that the frequency of model updates decreases as the anomaly probability threshold $T$ increases. For a fixed anomaly probability threshold $T$, the average execution time grows as the prediction horizon $L$ increases. This is due to the fact that increasing the prediction horizon $L$ makes it more difficult to satisfy the criteria for model updates.
\begin{figure}[!t]
\begin{center}
{\includegraphics[width=0.45\textwidth]{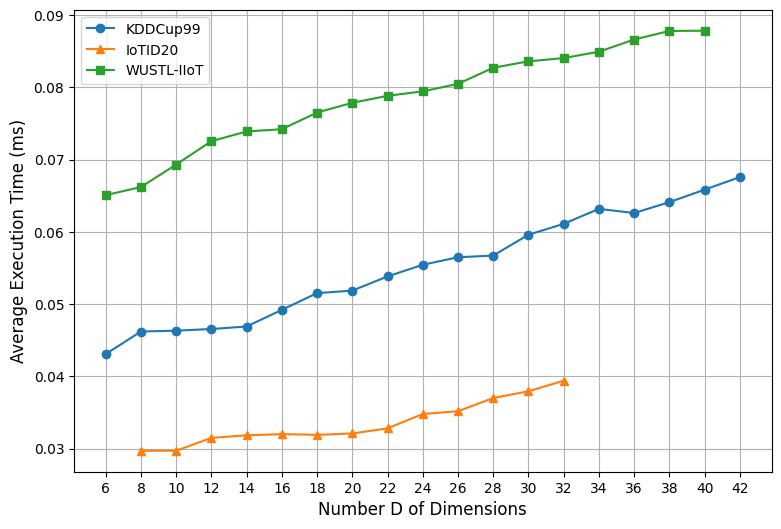}
}
\caption{\textcolor{black}{Average execution vs. number $D$ of dimensions.}}
\label{TimeDimsize}
\end{center}
\end{figure}
\paragraph{Scalability}

Next, we examine the scalability of the proposed algorithm. To ensure a fair comparison, Fig.~\ref{TimeDimsize} illustrates only the results of SAPDAD for different dimensions. The reason is that the execution time of the existing methods was well above a millisecond, even for a small number of dimensions. In the existing methods, as the number of dimensions increased, the average execution time increased significantly. \textcolor{black}{We vary the number $D$ of dimensions from 6 to 42, 8 to 32, and 6 to 40 for KDDCup99, IoTID20, and WUSTL-IIoT datasets, respectively}. Among the existing methods tested on the KDDCup99 dataset, the proposed SAPDAD method demonstrates the lowest execution time. More specifically, the obtained execution time is 1.2 ms and 3.6 ms for $D$=14 and $D$=42, respectively. Similarly, when applied to the IoTID20 dataset, the proposed SAPDAD method exhibits the lowest execution time of 1 ms and 2.8 ms for $D$=6 and $D$=32, respectively. \textcolor{black}{For the WUSTL-IIoT dataset, the proposed SAPDAD method exhibits the lowest execution time of 1.6 ms and 3.1 ms for $D$=24 and $D$=40, respectively.} We observe from Fig.~\ref{TimeDimsize} that the execution time of the proposed SAPDAD method grows linearly with the number of dimensions in the three datasets. The short execution time in the IoTID20 dataset compared to the KDDCup99 and WUSTL-IIoT datasets can be attributed to two factors. First, the network traffic patterns in the IoTID20 dataset have less variability and fluctuation, allowing for faster processing during testing. Second, given that there existed a smaller number of concept drifts in the IoTID20 dataset, fewer model adaptations were required during the testing.

\begin{table}[!t]
\centering

\caption{Contributors to the latency (ms) in the proposed SAPDAD.}
\label{CONTRIBUTORS}
\begin{tabular}{|c|c|c|c|c|}
\hline 
 phase & {Module} & \multicolumn{3}{c|}{Latency (ms)} \\ \cline{3-5}
 &  & KDDCup99 & IoTID20 & WUSTL-IIoT \\ 
\hline
{Offline} & Preprocessing & 0.091  &  0.050 & 0.12 \\ 
\cline{2-5} 
 & \makecell{Training and \\Prediction} & 3.110 & 2.544 & 3.857 \\ 
\hline
{\begin{tabular}[c]{@{}c@{}}Real-\\ Time\end{tabular}} & \makecell{Anomaly \\Detection} & 0.011  & 0.012 & 0.014 \\ 
\cline{2-5} 
 & \makecell{Drift Detection\\ and\\ Model Update} & 0.041  & 0.017 & 0.065 \\ 
\hline
\end{tabular}
\end{table}

\textcolor{black}{
\subsection{Latency of Various Modules in SAPDAD}
In order to identify the key factors contributing to latency in the proposed SAPDAD, a detailed breakdown of the latency for various modules is presented in Table~\ref{CONTRIBUTORS}. This table provides a comprehensive view of how each phase and module of the approach contributes to the overall latency across KDDCup99, IoTID20, and WUSTL-IIoT datasets.
We observe from Table~\ref{CONTRIBUTORS} that the Training and Prediction module is the primary contributor to the overall latency. This is particularly noticeable in the offline phase, where the latency of this module is significantly higher than the others, recording 3.110 ms for KDDCup99, 2.544 ms for IoTID20, and 3.857 ms for WUSTL-IIoT. The Preprocessing module, which is also in the offline phase, shows a latency of <0.1 ms for various datasets. This indicates a relatively minor impact on the overall latency, owing to the preprocessing algorithms' efficiency.
During the real-time phase, both Anomaly Detection and Drift Detection and Model Update modules demonstrate low latency values. The Anomaly Detection module records latencies of only 0.011 ms, 0.012 ms, and 0.014 ms for KDDCup99, IoTID20, and WUSTL-IIoT, respectively, reflecting its low computational complexity and efficient operation. Similarly, the Drift Detection and Model Update module shows latencies of 0.041 ms, 0.017 ms, and 0.065 ms for KDDCup99, IoTID20, and WUSTL-IIoT, respectively.}

\textcolor{black}{
\subsection{Ablation Study }
In the following, we run an ablation evaluation to examine the contribution of different modules of the proposed SAPDAD to the overall performance. 
}
\textcolor{black}{
\subsubsection{Impact of Drift Adaptation and Dimension Reduction}
Table~\ref{ablation} presents the impact of drift adaptation and dimension reduction modules of SAPDAD on the AUC performance across various datasets. The experiments involve four configurations: (i) full system with both modules enabled, (ii) system without drift adaptation, (iii) system without dimension reduction, and (iv) baseline system without any drift adaptation or dimension reduction. In the full system configuration, the obtained AUCs are the highest among all configurations, reaching 89.71\% for KDDCup99, 80.66\% for IoTID20, and 83.08\% for WUSTL-IIoT, which underscore the combined effectiveness of these modules in enhancing anomaly detection accuracy. When there is no drift adaptation, we observe a noticeable decline in AUC (85.75\% for KDDCup99, 76.64\% for IoTID20, and 79.10\% for WUSTL-IIoT), which highlights the importance of this feature in adapting to changes in data over time. Similarly, the absence of dimension reduction results in decreased AUCs of 87.75\%, 78.64\%, and 82.10\% for the respective datasets, indicating that managing the complexity of the feature space is crucial for maintaining high detection accuracy. The most significant impact is observed in the baseline configuration without any drift adaptation or dimension reduction, where the AUC drops to 82.75\%, 73.64\%, and 77.10\% for KDDCup99, IoTID20, and WUSTL-IIOT datasets, respectively.}

\begin{table}
\centering
\caption{Ablation experiment results for drift adaptation and dimension reduction on AUC in (A) KDDCup99, (B) IoTID20, and \textcolor{black}{(C) WUSTL-IIoT datasets.}}
\label{ablation}
\begin{tabular}{|l|c|c|c|c|c|} 
\hline
\multicolumn{1}{|c|}{{Variant}}                    & {\makecell{Drift \\ Adaptation}} &{\makecell{Dimension \\ Reduction}} & \multicolumn{3}{c|}{AUC (\%)}                     \\ 
\cline{4-6}
\multicolumn{1}{|l|}{}                                            & \multicolumn{1}{l|}{}                                                                                              & \multicolumn{1}{c|}{}                                                                                & A    & \multicolumn{1}{c|}{B} & \multicolumn{1}{c|}{C}  \\ 
\hline
Full System & Yes & Yes & 89.71 & 80.66 & 83.08 \\
\hline
\makecell{No Drift \\ Adaptation} & No & Yes & 85.75 & 76.64 &79.10  \\
\hline
\makecell{No Dimension \\ Reduction} & Yes & No & 87.75 & 78.64 &  82.10\\
\hline
No Modules & No & No & 82.75 & 73.64 & 77.10 \\
\hline
\end{tabular}
\end{table}
 
\begin{table}[!t]

\caption{Validation of recurring patterns on WUSTL-IIOT}
\label{ablation2}
   \begin{tabular}{|c|c|c|c|c|c|c|}
\hline Variant & $p_{1-\text { test }}$ & $p_{2-\text { test }}$ & $p_{3-\text { test }}$& $p_{4-\text { test }}$ &$p_{5-\text { test }}$ & \makecell{AUC\\ (\%)}\\
\hline \makecell{Variant \\I} & 82.0 & 79.0 & 77.5 & 76.0 & 81.0 & 79.1 \\
\hline \makecell{Variant \\II} & 82.0 & 70.0 & 68.5 & 67.0 & 72.0 & 71.9 \\
\hline
\end{tabular}
\end{table}

\textcolor{black}{
\subsubsection{Recurring Patterns} 
Table~\ref{ablation2} presents the results of two distinct training approaches used on the WUSTL-IIoT dataset, which was divided into five equal and sequential subsets, ranging from $p_1$ to $p_5$. Each subset was further split into training and test sets. For variant I, the GA-optimized hyperparameters from the $p_{1-\text{train}}$ subset are applied to all subsequent test subsets. We observe a slightly decreasing trend in AUC from $p_{1-\text{test}}$ to $p_{4-\text{test}}$, which is a direct consequence of concept drift. Specifically, as the subsets move further from the initial training data, the performance degrades due to changes in the underlying data distribution. However, the slight improvement in $p_{5-\text{test}}$ may suggest a recurring pattern or a return to similar conditions as the initial training set, allowing the model to regain some performance.
For variant II, the model is trained only once on $p_{1-\text{train}}$ and tested across all subsets. Here, a more significant performance drop is noted as we move to later subsets. The results for this variant can be explained by the lack of adaptation to concept drift. Given that no further optimization or retraining is conducted after the initial training set, the performance on the subsequent test sets degrades more dramatically. The use of GA for hyperparameter optimization on the initial training set should ideally provide robust hyperparameters that can generalize well to future data. However, as the data drifts, the optimized parameters may become less effective. Table~\ref{ablation2} reflects that optimized parameters perform well initially, but may require updating as the concept drift occurs. This is evidenced by the gradual performance decrease in both variants, more so in variant II, which does not adapt its hyperparameters.}

\begin{table}[!t]
\centering

\caption{Comparative Analysis of SAPDAD Model Performance using LSTM, CNN, GRU, and LSTM-CNN Architectures for WUSTL-IIoT.}
\label{models}
   \begin{tabular}{|l|c|c|}
\hline Model & AUC (\%) & Execution Time (ms) \\
\hline SAPDAD with LSTM & 83.08 & 0.0801 \\
\hline SAPDAD with CNN & 80.10 & 0.1176 \\
\hline SAPDAD with GRU & 79.95 & 0.6342 \\
\hline SAPDAD with LSTM-CNN  & 76.41 & 0.3452 \\
\hline
\end{tabular}
\end{table}

\textcolor{black}{\subsubsection{Performance Comparison of Various Architectures} 
Finally, we evaluate the accuracy of the trained prediction model of SAPDAD, which was initially implemented using LSTM. To explore the effectiveness of different neural network architectures, we replaced the LSTM network with three alternative configurations of CNN, GRU, and a hybrid network combining LSTM and CNN (LSTM-CNN). As shown in Table~\ref{models}, the LSTM configuration achieved the highest AUC of 84.10\%, with an execution time of 0.0801 ms. The CNN model, with an AUC of 80.10\%, exhibits a good balance between accuracy and efficiency, showing an execution time of 0.1176 ms, which may be suitable for environments with limited computational resources. The GRU variant presented a similar AUC of 79.95\%, but with a higher execution time of 0.6342 ms. In contrast, the LSTM-CNN hybrid, while offering extensive feature extraction capabilities, has a lower AUC of 76.41\% and a moderate execution time of 0.3452 ms, indicating the computational demands of integrating LSTM and CNN features for anomaly detection in IIoT.}


\section{Conclusions and Future Directions}
\label{conclusion}

The growing popularity of IIoT systems presents challenges in collecting and processing large volumes of streaming data generated by diverse sensors in dynamic environments. To address this challenge, adaptive machine learning methods are suitable solutions to deal with the continuously evolving IIoT data streams. In this paper, an adaptive model for real-time anomaly detection was proposed, which is accurate and time-efficient. The proposed SAPDAD method combines a novel drift-handling algorithm (RealTimeOAW), a prediction-driven algorithm (PDAD-SID), and a hyperparameter method (GA) to dynamically adapt to the continuous changes in data streams within IIoT environments. \textcolor{black}{The proposed SAPDAD method offers a number of advantages for anomaly detection in IIoT environments, including real-time processing capabilities, scalability, adaptability to data pattern changes, and improved accuracy with its prediction-driven strategy, which make it an effective solution for handling the dynamic and complex data streams in industrial environments.} \textcolor{black}{The evaluation of the proposed method was conducted using three real-world IIoT anomaly detection datasets, namely, KDDCup99, IoTID20, and WUSTL-IIoT.} Our trace-driven evaluations demonstrate that the proposed algorithm not only outperforms the existing benchmarks in terms of accuracy but can also detect anomalies in a bounded time. An interesting future research avenue is to develop a technique to consider time-sensitive application logs, resource metrics, and trace data to address the challenges associated with anomaly detection in large-scale complex microservice systems.

\section*{Acknowledgment}
This work is funded by the Ericsson/ENCQOR-5G Senior Industrial Research Chair on Cloud and Edge Computing for 5G and Beyond.

\bibliographystyle{ieeetr}
\bibliography{library}

@article{Dromard2017,
   author = {Juliette Dromard and Gilles Roudière and Philippe Owezarski},
   doi = {10.1109/TNSM.2016.2627340},
   issn = {19324537},
   issue = {1},
   journal = {IEEE Transactions on Network and Service Management},
   month = {3},
   pages = {34-47},
   publisher = {Institute of Electrical and Electronics Engineers Inc.},
   title = {Online and Scalable Unsupervised Network Anomaly Detection Method},
   volume = {14},
   year = {2017},
}

@article{Chouliaras2020,
   author = {Spyridon Chouliaras and Stelios Sotiriadis},
   issue = {9},
   journal = {IEEE Transactions on Industrial Informatics},
   pages = {6042-6049},
   title = {Real-Time Anomaly Detection of {NoSQL} Systems Based on Resource Usage Monitoring},
   volume = {16},
   year = {2020},
}

@article{Munir2019,
   author = {Mohsin Munir and Shoaib Ahmed Siddiqui and Andreas Dengel and Sheraz Ahmed},
   doi = {10.1109/ACCESS.2018.2886457},
   issn = {21693536},
   journal = {IEEE Access},
   pages = {1991-2005},
   publisher = {Institute of Electrical and Electronics Engineers Inc.},
   title = {{DeepAnT}: A Deep Learning Approach for Unsupervised Anomaly Detection in Time Series},
   volume = {7},
   year = {2019},
}

@article{Zhao2021,
   author = {Peihai Zhao and Xiaoyan Chang and Mimi Wang},
   doi = {10.1109/ACCESS.2021.3101844},
   issn = {21693536},
   journal = {IEEE Access},
   pages = {109025-109041},
   publisher = {Institute of Electrical and Electronics Engineers Inc.},
   title = {A novel multivariate time-series anomaly detection approach using an unsupervised deep neural network},
   volume = {9},
   year = {2021},
}

@article{Putina2021,
   author = {Andrian Putina and Dario Rossi},
   doi = {10.1109/TNSM.2020.3037019},
   issn = {19324537},
   issue = {1},
   journal = {IEEE Transactions on Network and Service Management},
   month = {3},
   pages = {839-854},
   publisher = {Institute of Electrical and Electronics Engineers Inc.},
   title = {Online Anomaly Detection Leveraging Stream-Based Clustering and Real-Time Telemetry},
   volume = {18},
   year = {2021},
}

@article{WuLSTM2020,
   author = {Di Wu and Zhongkai Jiang and Xiaofeng Xie and Xuetao Wei and Weiren Yu and Renfa Li},
   issue = {8},
   journal = {IEEE Transactions on Industrial Informatics},
   pages = {5244-5253},
   title = {{LSTM} Learning with Bayesian and Gaussian Processing for Anomaly Detection in Industrial {IoT}},
   volume = {16},
   year = {2020},
}

@article{Zhou2021,
   author = {Xiaokang Zhou and Yiyong Hu and Wei Liang and Jianhua Ma and Qun Jin},
   journal = {IEEE Transactions on Industrial Informatics},
   pages = {3469-3477},
   title = {Variational {LSTM} Enhanced Anomaly Detection for Industrial Big Data},
   volume = {17},
   year = {2021},
}

@inproceedings{laptev2015,
  title={Generic and scalable framework for automated time-series anomaly detection},
  author={Laptev, Nikolay and Amizadeh, Saeed and Flint, Ian},
  booktitle={Proc. ACM SIGKDD International Conference on Knowledge Discovery and Data Mining},
  pages={1939--1947},
  year={2015}
}

@inproceedings{liu2015,
title={Opprentice: Towards practical and automatic anomaly detection through machine learning},
author={Liu, Dapeng and Zhao, Youjian and Xu, Haowen and Sun, Yongqian and Pei, Dan and Luo, Jiao and Jing, Xiaowei and Feng, Mei},
  booktitle={Proc. ACM Internet Measurement Conference},
  pages={211--224},
  year={2015}
}

@inproceedings{seabold2010statsmodels,
  title={Statsmodels: Econometric and statistical modeling with {P}ython},
  author={Seabold, Skipper and Perktold, Josef},
  booktitle={Proc. 9th Python in Science Conference},
  year={2010},
}

@inproceedings{malhotra2015,
  title={Long short term memory networks for anomaly detection in time series},
  author={Malhotra, Pankaj and Vig, Lovekesh and Shroff, Gautam and Agarwal, Puneet and others},
  booktitle={Proc. European Symposium on Artificial Neural Networks (ESANN)},
  volume={89},
  pages={89--94},
  year={2015}
}

@article{ahmad2017,
  title={Unsupervised real-time anomaly detection for streaming data},
  author={Ahmad, Subutai and Lavin, Alexander and Purdy, Scott and Agha, Zuha},
  journal={Elsevier Neurocomputing},
  volume={262},
  pages={134--147},
  year={2017},
}

@inproceedings{spinosa2008cluster,
  title={Cluster-based novel concept detection in data streams applied to intrusion detection in computer networks},
  author={Spinosa, Eduardo J and de Leon F. de Carvalho, Andr{\'e} Ponce and Gama, Jo{\~a}o},
  booktitle={Proc. {ACM} Symposium on Applied computing},
  pages={976--980},
  year={2008}
}

@inproceedings{saurav2018online,
  title={Online anomaly detection with concept drift adaptation using recurrent neural networks},
  author={Saurav, Sakti and Malhotra, Pankaj and TV, Vishnu and Gugulothu, Narendhar and Vig, Lovekesh and Agarwal, Puneet and Shroff, Gautam},
  booktitle={Proc. {ACM} india joint international conference on data science and management of data},
  pages={78--87},
  year={2018}
}

@inproceedings{ullah2020scheme,
  title={A scheme for generating a dataset for anomalous activity detection in {IoT} networks},
  author={Ullah, Imtiaz and Mahmoud, Qusay H},
  booktitle={Proc. Canadian Conference on Artificial Intelligence},
  pages={508--520},
  year={2020},
  organization={Springer}
}

@article{wang2021concept,
  title={Concept Drift-based Runtime Reliability Anomaly Detection for Edge Services Adaptation},
  author={Wang, Lei and Chen, Shuhan and He, Qiang},
  journal={IEEE Transactions on Knowledge and Data Engineering},
  year={2021},
  publisher={IEEE}
}

@article{van1997,
  title={{SUMMA}: Scalable universal matrix multiplication algorithm},
  author={Van De Geijn, Robert A and Watts, Jerrell},
  journal={Concurrency: Practice and Experience},
  volume={9},
  number={4},
  pages={255--274},
  year={1997},
  publisher={Wiley Online Library}
}

@article{yang2022astream,
  title={{ASTREAM}: Data-Stream-Driven Scalable Anomaly Detection with Accuracy Guarantee in {IIoT} Environment},
  author={Yang, Yihong and Yang, Xuan and Heidari, Mohsen and Khan, Mohammad AYOUB and Srivastava, Gautam and Khosravi, Mohammad and Qi, Lianyong},
  journal={IEEE Transactions on Network Science and Engineering},
   year={2023},
  volume={10},
  number={5},
  pages={3007-3016},
}

@article{jiang2021road,
  title={The road towards {6G}: A comprehensive survey},
  author={Jiang, Wei and Han, Bin and Habibi, Mohammad Asif and Schotten, Hans Dieter},
  journal={IEEE Open Journal of the Communications Society},
  volume={2},
  pages={334--366},
  year={2021},
  publisher={IEEE}
}

@article{ippoliti2016online,
  title={Online adaptive anomaly detection for augmented network flows},
  author={Ippoliti, Dennis and Jiang, Changjun and Ding, Zhijun and Zhou, Xiaobo},
  journal={ACM Transactions on Autonomous and Adaptive Systems (TAAS)},
  volume={11},
  number={3},
  pages={1--28},
  year={2016},
  publisher={ACM New York, NY, USA}
}

@article{yu2015scalable,
  title={A scalable, non-parametric method for detecting performance anomaly in large scale computing},
  author={Yu, Li and Lan, Zhiling},
  journal={IEEE Transactions on Parallel and Distributed Systems},
  volume={27},
  number={7},
  pages={1902--1914},
  year={2015},
  publisher={IEEE}
}

@article{yang2020hyperparameter,
  title={On hyperparameter optimization of machine learning algorithms: Theory and practice},
  author={Yang, Li and Shami, Abdallah},
  journal={Neurocomputing},
  volume={415},
  pages={295--316},
  year={2020},
  publisher={Elsevier}
}

@article{dealmeida2021abnormal,
  title={Abnormal Behavior Detection Based on Traffic Pattern Categorization in Mobile Networks},
  author={DeAlmeida, Jonathan M and Pontes, Camila FT and DaSilva, Luiz A and Both, Cristiano B and Gondim, Jo{\~a}o JC and Ralha, Celia G and Marotta, Marcelo A},
  journal={IEEE Transactions on Network and Service Management},
  volume={18},
  number={4},
  pages={4213--4224},
  year={2021},
  publisher={IEEE}
}

@inproceedings{raeiszadeh2023,
  title={A Deep Learning Approach for Real-Time Application-Level Anomaly Detection in {IoT} Data Streaming},
  author={Raeiszadeh, Mahsa and Saleem, Ahsan and Ebrahimzadeh, Amin and Glitho, Roch H and Eker, Johan and Mini, Raquel AF},
  booktitle={Proc. IEEE Consumer Communications \& Networking Conference (CCNC)},
  pages={449--454},
  year={2023},
  
}

@article{yang2022iot,
  title={{IoT} data analytics in dynamic environments: From an automated machine learning perspective},
  author={Yang, Li and Shami, Abdallah},
  journal={Engineering Applications of Artificial Intelligence},
  volume={116},
  pages={105366},
  year={2022},
  publisher={Elsevier}
}

@article{nassif2021machine,
  title={Machine learning for anomaly detection: A systematic review},
  author={Nassif, Ali Bou and Talib, Manar Abu and Nasir, Qassim and Dakalbab, Fatima Mohamad},
  journal={IEEE Access},
  volume={9},
  pages={78658--78700},
  year={2021},
  publisher={IEEE}
}

@article{abdi2010principal,
  title={Principal component analysis},
  author={Abdi, Herv{\'e} and Williams, Lynne J},
  journal={Wiley interdisciplinary reviews: computational statistics},
  volume={2},
  number={4},
  pages={433--459},
  year={2010},
  publisher={Wiley Online Library}
}

@article{wang2022improved,
  title={Improved {LSTM}-based time-series anomaly detection in rail transit operation environments},
  author={Wang, Yujie and Du, Xin and Lu, Zhihui and Duan, Qiang and Wu, Jie},
  journal={IEEE Transactions on Industrial Informatics},
  volume={18},
  number={12},
  pages={9027--9036},
  year={2022},
  publisher={IEEE}
}

@article{hao2021hybrid,
  title={Hybrid statistical-machine learning for real-time anomaly detection in industrial cyber-physical systems},
  author={Hao, Weijie and Yang, Tao and Yang, Qiang},
  journal={IEEE Transactions on Automation Science and Engineering},
  year={2021},
  publisher={IEEE}
}

@article{xie2022abnormal,
  title={Abnormal Data Detection Based on Adaptive Sliding Window and Weighted Multiscale Local Outlier Factor for Machinery Health Monitoring},
  author={Xie, Qinglin and Tao, Gongquan and Xie, Chenxi and Wen, Zefeng},
  journal={IEEE Transactions on Industrial Electronics},
  year={2022},
  publisher={IEEE}
}

@article{li2021similarity,
  title={Similarity-measured isolation forest: Anomaly detection method for machine monitoring data},
  author={Li, Changgen and Guo, Liang and Gao, Hongli and Li, Yi},
  journal={IEEE Transactions on Instrumentation and Measurement},
  volume={70},
  pages={1--12},
  year={2021},
  publisher={IEEE}
}

@article{quek2019iot,
  title={{IoT} load classification and anomaly warning in ELV DC picogrids using hierarchical extended k-nearest neighbors},
  author={Quek, Yang Thee and Woo, Wai Lok and Thillainathan, Logenthiran},
  journal={IEEE Internet of Things Journal},
  volume={7},
  number={2},
  pages={863--873},
  year={2021},
  publisher={IEEE}
}

@article{paulauskas2019application,
  title={Application of histogram-based outlier scores to detect computer network anomalies},
  author={Paulauskas, Nerijus and Baskys, Algirdas},
  journal={Electronics},
  volume={8},
  number={11},
  pages={1251},
  year={2019},
  publisher={MDPI}
}

@misc{zolanvari2019machine,
  author = {Zolanvari, M. and Teixeira, M. A. and Gupta, L. and Khan, K. M. and Jain, R.},
  title = {{WUSTL-IIOT}-2021 Dataset for {IIoT} Cybersecurity Research},
  howpublished = {Washington University in St. Louis, USA},
  month = oct,
  year = {2021\color{black}},
  url = {http://www.cse.wustl.edu/~jain/iiot2/index.html}
}

@article{mothukuri2021federated,
  title={Federated-learning-based anomaly detection for {IoT} security attacks},
  author={Mothukuri, Viraaji and Khare, Prachi and Parizi, Reza M and Pouriyeh, Seyedamin and Dehghantanha, Ali and Srivastava, Gautam},
  journal={IEEE Internet of Things Journal},
  volume={9},
  number={4},
  pages={2545--2554},
  year={2021\color{black}},
  publisher={IEEE}
}

@article{aman2022machine,
  title={Machine-Learning-Based Attestation for the Internet of Things Using Memory Traces},
  author={Aman, Muhammad Naveed and Basheer, Haroon and Wong, Jun Wen and Xu, Jia and Lim, Hoon Wei and Sikdar, Biplab},
  journal={IEEE Internet of Things Journal},
  volume={9},
  number={20},
  pages={20431--20443},
  year={2022\color{black}},
  publisher={IEEE}
}

@inproceedings{javaid2020defining,
  title={Defining trust in {IoT} environments via distributed remote attestation using blockchain},
  author={Javaid, Uzair and Aman, Muhammad Naveed and Sikdar, Biplab},
  booktitle={Proc. ACM International Symposium on Theory, Algorithmic Foundations, and Protocol Design for Mobile Networks and Mobile Computing},
  pages={321--326},
  year={2020\color{black}}
}

@inproceedings{banik2023anomaly,
  title={Anomaly Detection Techniques in Smart Grid Systems: A Review},
  author={Banik, Shampa and Saha, Sohag Kumar and Banik, Trapa and Hossain, SM Mostaq},
  booktitle={Proc. IEEE World AI IoT Congress (AIIoT)},
  pages={0331--0337},
  year={2023\color{black}}

}

@inproceedings{domenech2016providing,
  title={Providing a smart industrial environment with the web of things and cloud computing},
  author={Domenech, Marlon C and Rauta, Leonardo P and Lopes, Marcelo Dornbusch and Da Silva, Paulo H and Da Silva, Rodrigo C and Mezger, Benjamin W and Wangham, Michelle S},
  booktitle={Proc. IEEE International Conference on Services Computing (SCC)},
  pages={641--648},
  year={2016},
}

@article{nandanwar2024deep,
  author={Nandanwar, Himanshu and Katarya, Rahul},
  title={Deep learning enabled intrusion detection system for Industrial {IoT} environment},
  journal={Expert Systems with Applications},
  volume={249},
  pages={123808},
 year={2024},
  publisher={Elsevier}
}

@inproceedings{kauhsik2023iot,
 author={Kauhsik, Bhanu and Nandanwar, Himanshu and Katarya, Rahul},
  title={{IoT} Security: A Deep Learning-Based Approach for Intrusion Detection and Prevention},
  booktitle={Proc. IEEE International Conference on Evolutionary Algorithms and Soft Computing Techniques (EASCT)},
  pages={1--7},
  year={2023},
}

\vspace{-0.3cm}
\begin{IEEEbiography}[{\includegraphics[width=1in,height=1.25in,clip,keepaspectratio]{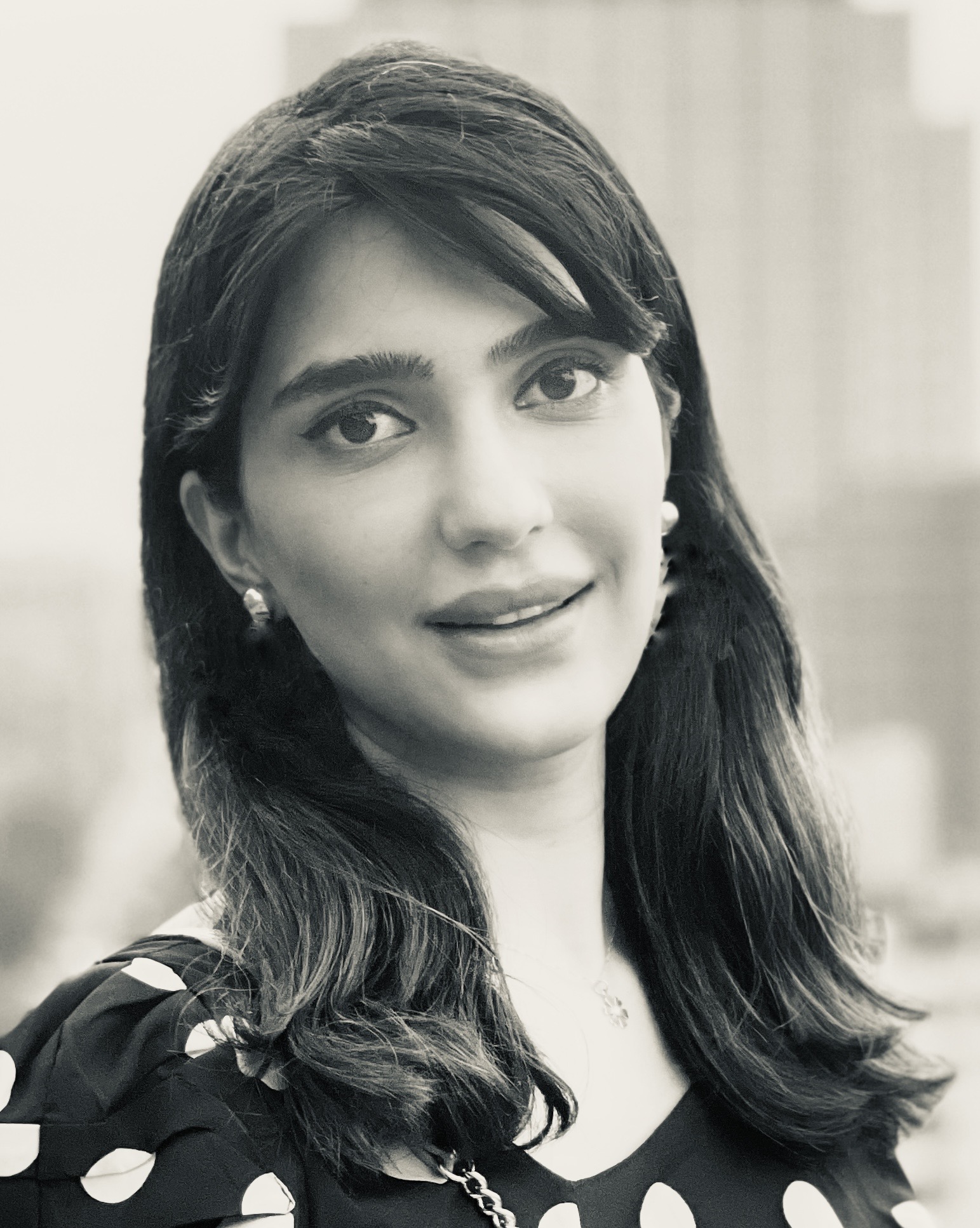}}]{Mahsa Raeiszadeh}
received the B.Sc. and M.Sc.
degrees in IT engineering from Shiraz University of Technology and Shiraz University, Shiraz,
Iran, in 2014 and 2018, respectively. From 2018 to 2020, she was with Shiraz University, Shiraz, Iran. She is currently pursuing a Ph.D. degree and is a member of
Telecommunication Service Engineering Research Lab, Concordia University, Montreal, QC, Canada. Her research interests include cloud computing, IoT, machine learning, and anomaly detection.
\end{IEEEbiography}
\vspace{-0.3cm}
\begin{IEEEbiography}[{\includegraphics[width=1in,height=1.25in,clip,keepaspectratio]{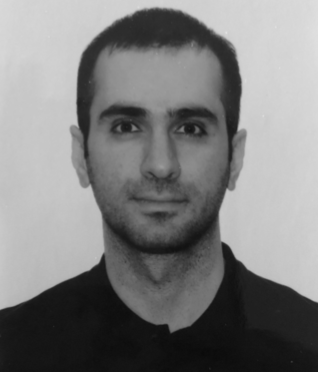}}]{Amin Ebrahimzadeh}
received the B.Sc. and M.Sc.
degrees in electrical engineering from the University of
Tabriz, Tabriz, Iran, in 2009 and 2011, respectively,
and the Ph.D. degree (Hons.) in telecommunications
from the Institut National de la Recherche Scientifique,
Montreal, QC, Canada, in 2019. From 2011 to 2015, he
was with the Sahand University of Technology, Tabriz,
Iran. He is currently a Horizon Postdoctoral Fellow
with Concordia University, Montreal, QC, Canada. His
research interests include 6G networks, tactile internet,
FiWi networks, and multiaccess edge computing. He
has been awarded the 2019–2020 Best Doctoral Thesis Prize of INRS for his
research on the tactile internet. He was the recipient of the doctoral research
scholarship from the B2X program of Fonds de Recherche du Quebec-Nature et
Technologies. He is the Lead Author of the book \textit{Toward 6G: A New Era of Convergence} (Wiley-IEEE Press, 2021)
\end{IEEEbiography}
\vspace{-0.3cm}
\begin{IEEEbiography}[{\includegraphics[width=1in,height=1.25in,clip,keepaspectratio]{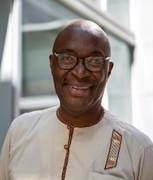}}]{Roch H. Glitho}
(Senior Member, IEEE) received
the M.Sc. degree in business economics from
the University of Grenoble, France, the M.Sc.
degree in pure mathematics and the M.Sc. degree
in computer science from the University of
Geneva, Switzerland, and the Ph.D. (Tech.Dr.)
degree in informatics from the Royal Institute of
Technology, Stockholm, Sweden. He is currently a
Full Professor with Concordia University, where he
holds a Canada Research Chair. He also holds the
Ericsson/ENCQOR-5G Senior Industrial Research
Chair in Cloud and Edge Computing for 5G and Beyond. In addition, he is a
Professor extraordinaire with the Computer Science Department, University of
the Western Cape, Cape Town, South Africa. He has worked in industry and
has held several senior technical positions (e.g., a Senior Specialist, Principal
Engineer, and Expert) with Ericsson, Sweden and Canada. He has also
served as an IEEE Distinguished Lecturer, and the Editor-in-Chief of IEEE
Communications Magazine and the IEEE COMMUNICATIONS SURVEYS \& TUTORIALS.
\end{IEEEbiography}
\vspace{-0.3cm}
\begin{IEEEbiography}[{\includegraphics[width=1in,height=1.25in,clip,keepaspectratio]{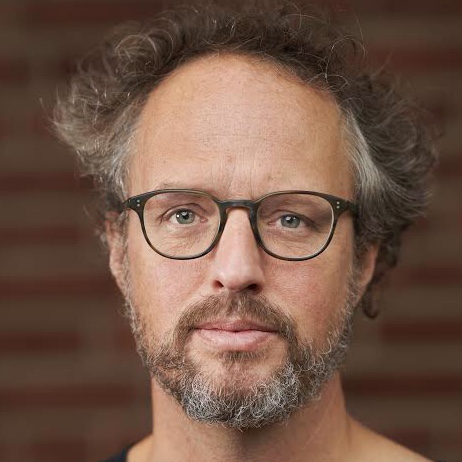}}]{Johan Eker} is a Principal Researcher at Ericsson Research, Sweden. He received his Ph.D. in Automatic Control from Lund University in 1999 and subsequently joined the Ptolemy group at UC Berkeley. Since 2023 he is also a Professor in Automatic Control at Lund University with focus on real-time systems. He is the head of the WASP research arena on data-driven operations (WARA-Ops). His current research focus is on control of large-scale compute systems, but his research interests range from programming language design for parallel hardware, real-time control systems, mobile communications. software design for mobile devices, adaptive resource management, IoT and cloud technology. He is the co-designer of the CAL Actor Language, which is part of the MPEG standard ISO/IEC 23001-4:2011. He holds over 70 granted patents in the areas of telecom, IoT and cloud computing. He is participating in a range of program committees and research projects on topics such as real-systems, signal processing, software development, cloud technology, brain-computer interfaces, and AI. He is involved in the operation of the Ericsson Research Data Center and works with industrial cloud applications.
\end{IEEEbiography}
\vspace{-0.3cm}
\begin{IEEEbiography}[{\includegraphics[width=1in,height=1.25in,clip,keepaspectratio]{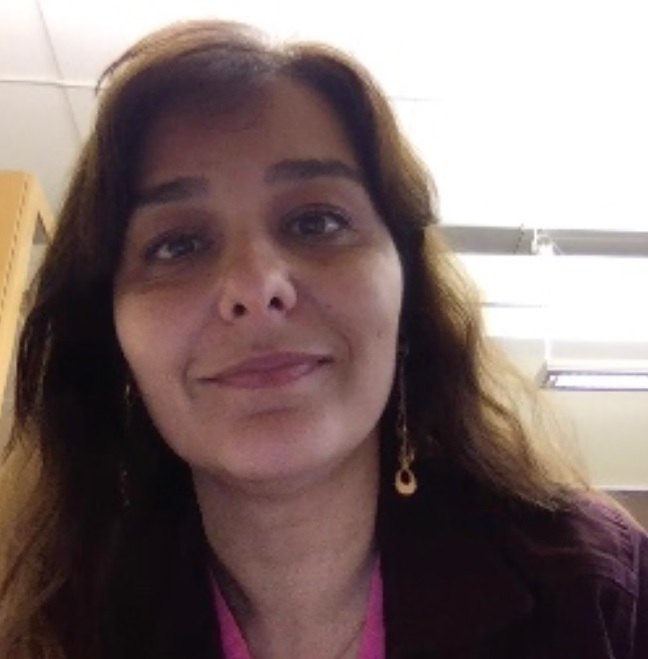}}]{Raquel A. F. Mini} is a researcher at Ericsson working in the Cloud Systems and Platforms area. She got her BSc, MSc, and PhD in Computer Science from Federal University of Minas Gerais (UFMG), Brazil. Before joining Ericsson in 2021, Raquel worked for more than 20 years as a Professor of Computer Science in Brazil. Her research addresses the area of sensor networks, IoT, ubiquitous computing, and cloud computing.

\end{IEEEbiography}
 \vfill

\end{document}